\newif\ifkouetsu
\kouetsufalse
\newif\ifannotation
\annotationfalse
\ifkouetsu
\documentclass[a4paper,12pt]{article}
\else
\documentclass[paper]{./ieice}
\fi

\newcommand{\ignore}[1]{}
\newcommand{\mycom}[1]{}
\usepackage[pdftex]{graphicx,xcolor}
\usepackage[fleqn]{empheq}
\usepackage{bm}
\usepackage{amsxtra}
\usepackage{multirow}
\usepackage{cite}
\usepackage{newtxtext}
\usepackage{subfigure}

\ifkouetsu

\else

\fi
\usepackage[varg]{newtxmath}
\usepackage{color}
\usepackage{soul}
\ifannotation
\newcommand{\myhl}[2]{\MNOTE[red]{\textcolor{red}{#2}}\!\!\hl{#1}}
\newcommand{\myhlf}[2]{\hl{#1}\ignore{#2}}
\newcommand*\myybox[1]{%
  \colorbox{yellow}{\hspace{0em}#1\hspace{0em}}}
\newenvironment{ybox}{%
  \empheq[box=\myybox]{align}}{\endempheq}
\else
\newcommand{\myhl}[2]{#1\ignore{#2}}
\newcommand{\myhlf}[2]{#1\ignore{#2}}
\newcommand*\myybox[1]{%
  \colorbox{yellow}{\hspace{0em}#1\hspace{0em}}}
\newenvironment{ybox}{%
  \align}{\endalign}
  \fi
\begin{document}
\title{%
Two-Layer Lossless HDR Coding using Histogram Packing Technique with Backward Compatibility to JPEG
}%
\ifkouetsu
\date{}
\else
\authorlist{
\authorentry[owatanab@es.takushoku-u.ac.jp]{Osamu WATANABE}{m}{Tack}
\authorentry[hkob@metro-cit.ac.jp]{Hiroyuki KOBAYASHI}{m}{cit}
\authorentry[kiya@tmu.ac.jp]{Hitoshi KIYA}{f}{tmu}
\affiliate[Tack]{Dept. of Electronics \& Computer Systems, Takushoku
University.}
\affiliate[cit]{Tokyo Metropolitan College of Industrial Technology}
\affiliate[tmu]{Faculty of Info. and Commun. Systems, Tokyo Metropolitan University}
}
\received{2018}{1}{1}
\revised{2018}{1}{1}
\field{A}
\SpecialSection{Smart Multimedia \& Communication Systems}
\fi
\maketitle
 \begin{summary}
 An efficient two-layer coding method using the histogram packing
 technique with the backward compatibility to the legacy JPEG is proposed
 in this paper. The JPEG XT, which is the international standard to
 compress HDR images, adopts two-layer coding scheme for backward
 compatibility to the legacy JPEG. However, this two-layer coding structure does not
 give better lossless performance than the other existing methods for HDR
 image compression with single-layer structure. Moreover, 
 the lossless compression of the JPEG XT has \myhl{a problem}{B1} on
 determination of the coding parameters;
 The lossless performance is affected by the input images and/or the
 parameter values. That is, finding appropriate
 combination of the values is necessary to achieve good lossless
 performance.
\myhl{It is firstly pointed out that the histogram packing technique
considering the histogram sparseness of HDR images is able to improve
the performance of lossless compression. Then, a novel two-layer coding
 with the histogram packing technique and an additional lossless
 encoder is proposed.}{B2,D1}
The experimental results demonstrate that not only the proposed method
 has a better lossless compression
 performance than that of the JPEG XT, but also there is no need to determine
image-dependent parameter values for good compression
 performance \myhl{without losing}{D2}
the backward
 compatibility to the well known legacy JPEG standard.
 \end{summary}
\section{Introduction}
The image compression method designed to provide coded data containing
high dynamic range content is highly expected to meet the rapid growth
of high dynamic range (HDR) image applications. Generally, HDR images
have much greater bit depth of pixel values and much wider color gamut\cite{ReinhardBook,8026195,ArtusiBook01,BADC11}.
These characteristic of HDR images are suitable for many applications,
such as cinema, medical and masterpieces of art etc.
For such applications, HDR images should be often losslessly encoded. In other
words, \myhl{they should be compressed without any coding loss.}{D3}

Most of conventional image compression methods, however, could not
efficiently compress HDR image due to its greater bit depth and uncommon
pixel format including a floating-point based pixel encoding.
\myhl{For example, in very
limited numbers of HDR rendering applications, RGBE pixel format is
used. This format stores pixels for RGB in the same manner of the
existing representation of uncompressed RGB data with a one byte shared exponent for floating-point representation.}{D4}
Several methods have been proposed for compression of HDR
images\cite{Ward:2006:JBH:1185657.1185685,1528434,doi:10.1117/12.805315,6622714,4517823,6287996,6411962,6637869,7991151} and 
ISO/IEC JTC 1/SC 29/WG 1
\myhl{(JPEG) has developed a series of international standards}{D5} referred to as JPEG
XT\cite{ThomasPCS2013,Artusi2015,7426553,JPEGXT,7535096} \myhl{for compression of 
HDR images.}{D5} The JPEG XT has been designed to be backward compatible with
the legacy JPEG\cite{JPEG-1} with two-layer coding; a base layer for
tone-mapped LDR image is compressed by the legacy JPEG encoder and an extension layer for residual data
consists of the subtracted data between a partially decoded base layer image
and an original HDR image is compressed \myhl{by the JPEG-based encoding
procedure which consists of color conversion, scalar quantization with
tables and huffman coding.}{D6} This
backward compatibility to the legacy JPEG allows legacy applications and
existing toolchains to continue to operate on codestreams conforming to
JPEG XT. Besides this two-layer coding procedure makes it possible to
compress HDR images with the backward compatibility and the extension
layer contributes the improvement of the decoded image quality in lossy
compression\cite{7532739},
lossless compression is possible by using the residual data in the
extension layer.
\newsavebox{\boxb}\sbox{\boxb}{\cite{JPEGXTpart8}}%
\myhl{The ISO/IEC IS 18477-8:2016{\usebox\boxb}}{D7}, which is known as the \myhl{JPEG XT
Part 8}{D27}, \myhl{describes how to decode losslessly or near-losslessly encoded HDR images.}{D7}
In this \myhl{Part 8}{D27},
its lossless compression
performance is not better than that of the other existing methods for HDR image
compression with single coding layer procedure and
it is required to find a combination of the parameter
values which gives a good lossless compression performance. The
combination could be dependent on input HDR images. That is, finding the
combination is required to compress HDR images losslessly and
efficiently.
In Refs.\cite{6288143,Minewaki2017,958146,1034993,988715,1040040,ELCVIA116,eusipco2017,6411962,6213328,6637869}, the sparseness of a histogram of an image is used
for efficient compression. \myhl{`Sparse'}{D28} histogram means that not all the bins in
a histogram are utilized. It is well known that a histogram of an HDR
image shows a tendency to be sparse\cite{6411962,6637869}.
 In Refs.\cite{6637869,KIYAJan2018}, methods for two-layer lossless coding 
of HDR images have been proposed, however, they are not backward
compatible with the legacy JPEG.

\begin{figure*}[tb]
 \centering
 \ifannotation
 \includegraphics[width=1.35\columnwidth]{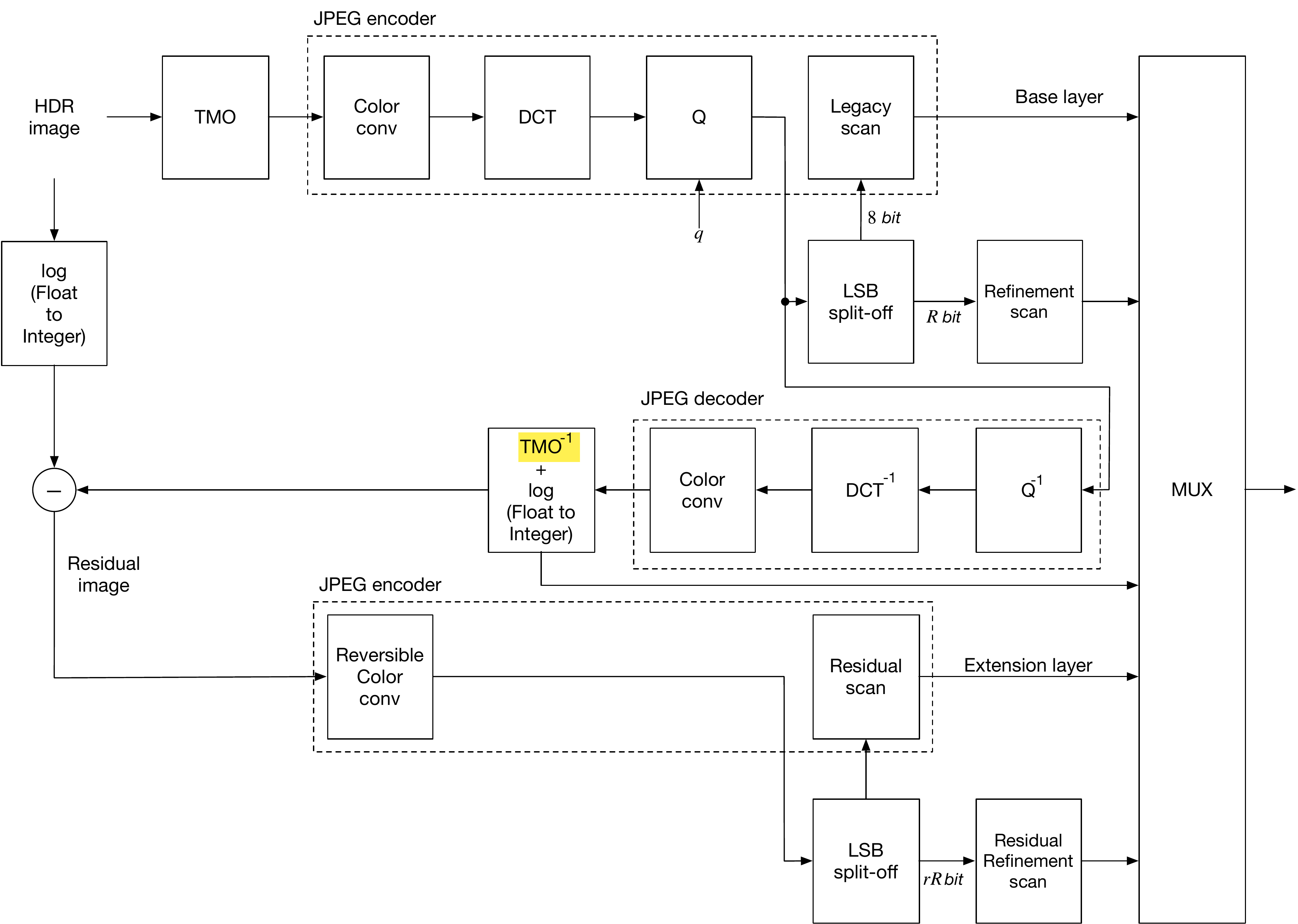}
 \else
 \includegraphics[width=1.35\columnwidth]{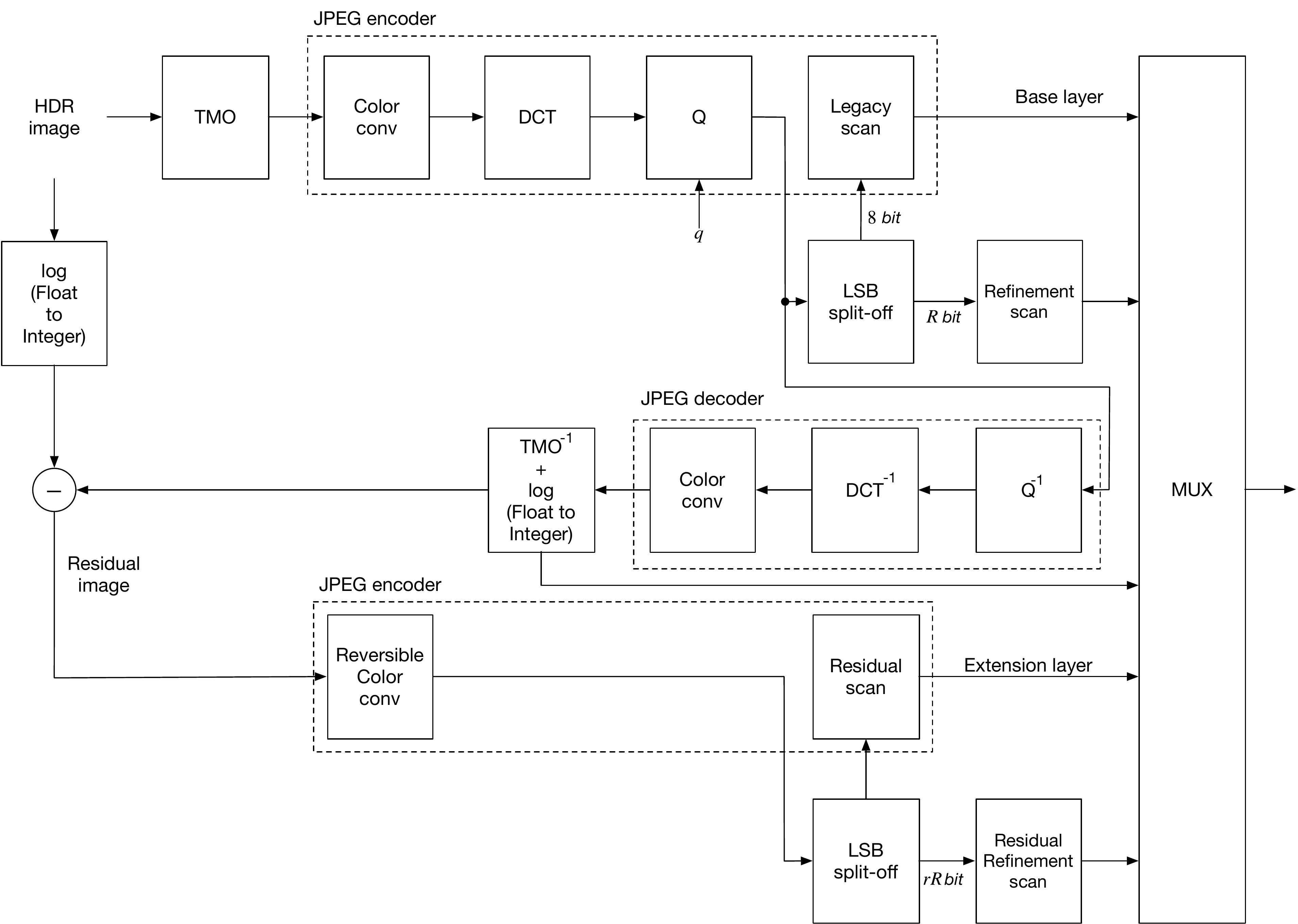}
 \fi
 \caption{Blockdiagram of \myhlf{JPEG XT Part 8}{D27} encoder: \myhlf{`TMO'}{D28} means tone
 mapping operator. $Q$ and $Q^{-1}$ are quantization and inverse
 quantization, respectively. $q$ is parameter to control quality of
 decoded base layer (LDR) image.}
 \label{XT_C_enc}
\end{figure*}
\begin{figure}[tb]
 \centering
 \includegraphics[width=1.0\columnwidth]{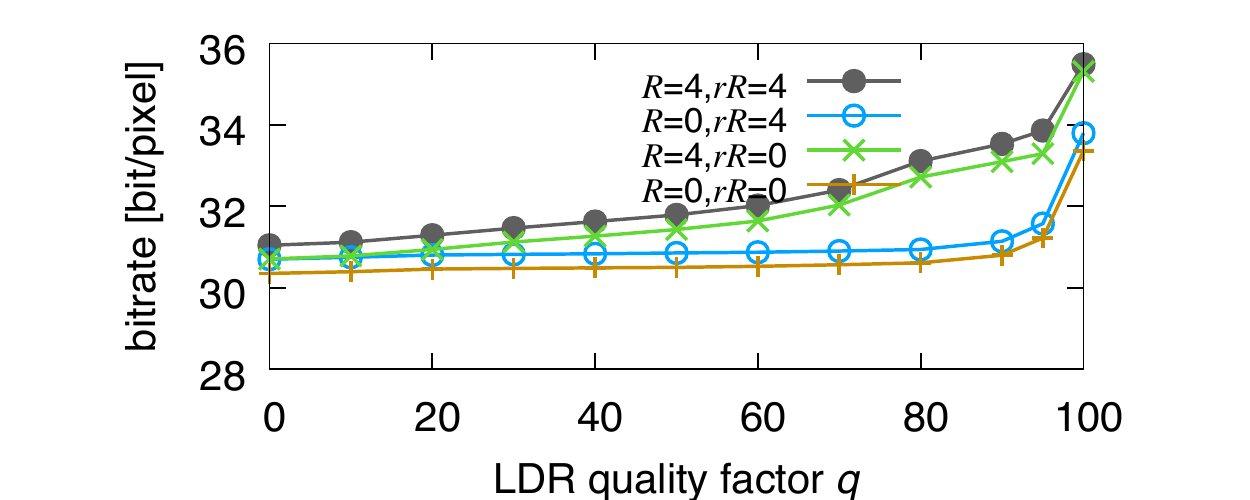}
 \caption{Bitrate of lossless compressed HDR image (MtTamWest) by \myhlf{JPEG
 XT Part 8}{D27}}
 \label{XTwithRef}
\end{figure}

This paper proposes a
new lossless two-layer method for both integer and floating-point HDR
images with the histogram packing technique.
Codestreams
produced by the proposed method consist of two layers, i.e.
base layer and extension layer, where the base layer provides
low dynamic range (LDR) images mapped from HDR images by a
tone mapping operator (TMO), while the extension layer has the residual
information for reconstructing the original HDR images.
For those residual data, any lossless image encoders that can handle
over \myhl{16 bits}{D8}, such as JPEG 2000 and JPEG XR, could be used.
In
addition, the codestreams for the base layer are compatible with legacy JPEG
decoders.
%
%
Not only the proposed method has a higher compression
performance than that of the \myhl{JPEG XT Part 8}{D27},
%
%
but also there is no need to determine
image-dependent parameter values to achieve good compression performance,
\myhl{because no coding parameter exists to compress the residual data for the extension layer.}{D9}
%

 \section{\myhl{Problem}{B1} with \myhl{JPEG XT Part 8}{D27}}
\myhl{In this section, the coding procedure of the JPEG XT Part 8 is
 summarized and then the problem with it is described.}{D10,D27}
\ignore{Because we
focus the lossless coding of HDR images with backward compatibility to
the legacy JPEG decoders,
the coding procedure of the \myhl{Part 8}{D27} of JPEG XT is summarized and then
the problem with it is described in this section.}

%
The blockdiagram of
the \myhl{Part 8}{D27} encoder is shown in Fig.\ref{XT_C_enc}.
Although the pixel values of HDR images are often represented with
floating-point numbers, these floating-point numbers are re-interpreted as
integer number with IEEE floating-point
representation\cite{4610935,595279}. This
representation is exactly invertible\cite{ThomasXT} and makes it
possible to compress HDR images losslessly.
For lossless compression of HDR images, it is required to determine the
values of several parameters. The first parameter is $q$, which
controls decoded image quality of base layer. The higher $q$ gives the
better quality. The second parameter $R$ is the number of bits used for
refinement scan. The refinement scan is used to improve precision of DCT
coefficients up to 12 bit. Thus the valid range of $R$ is from $0$ to $4$.
The third parameter is $rR$. The $rR$ is the number of bits used for
the residual refinement scan. In this lossless coding procedure, the $rR$
is considered as the control factor for the amount of coded data
included in the residual data of the extension layer.

To achieve good lossless compression performance, the values of the
parameters, $q$, $R$ and $rR$ should be carefully determined. Figure
\ref{XTwithRef} shows the result of lossless compression of an HDR image
 with from $q=0$ to $q=100$, $R=4$ and $rR=0$.
\newsavebox{\boxh}\sbox{\boxh}{\ref{comp_xt_pro}}
 \myhl{The other examples with
 the different combination of the parameter values are depicted in Fig. {\usebox\boxh}.}{D12}

 Clearly, we can see there is a certain variation in the coding
performance. Note that it has been confirmed that the optimal values of the
parameters which give the best performance is image-dependent.

\section{Proposed method}
A method using the histogram packing technique with
the two-layer coding having the backward compatibility to the legacy
JPEG for base layer is described in this section.
  \subsection{Histogram sparseness of residual data}
\label{IIB}
\begin{figure}[tb]
 \centering
 \newcommand{\histW}{0.75}
\subfigure[memorial]{
 \includegraphics[width=\histW\columnwidth]{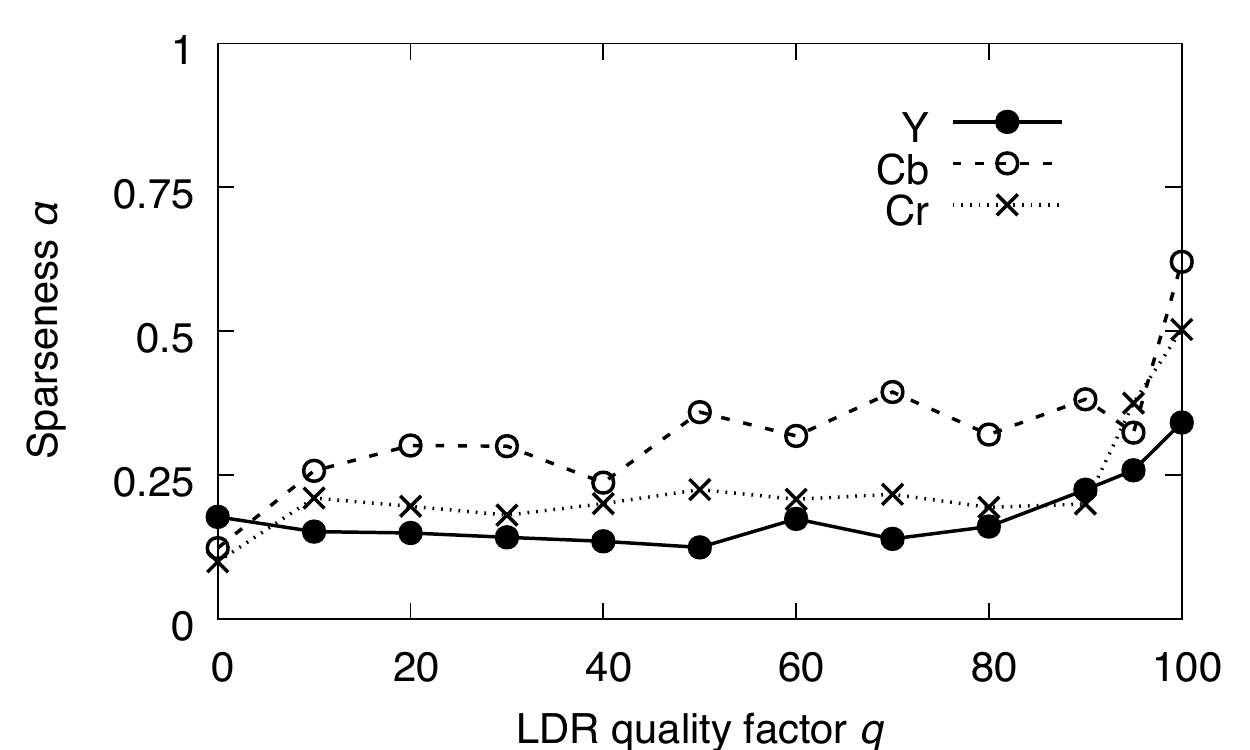}
 }
 \subfigure[Blooming Gorse2]{
 \includegraphics[width=\histW\columnwidth]{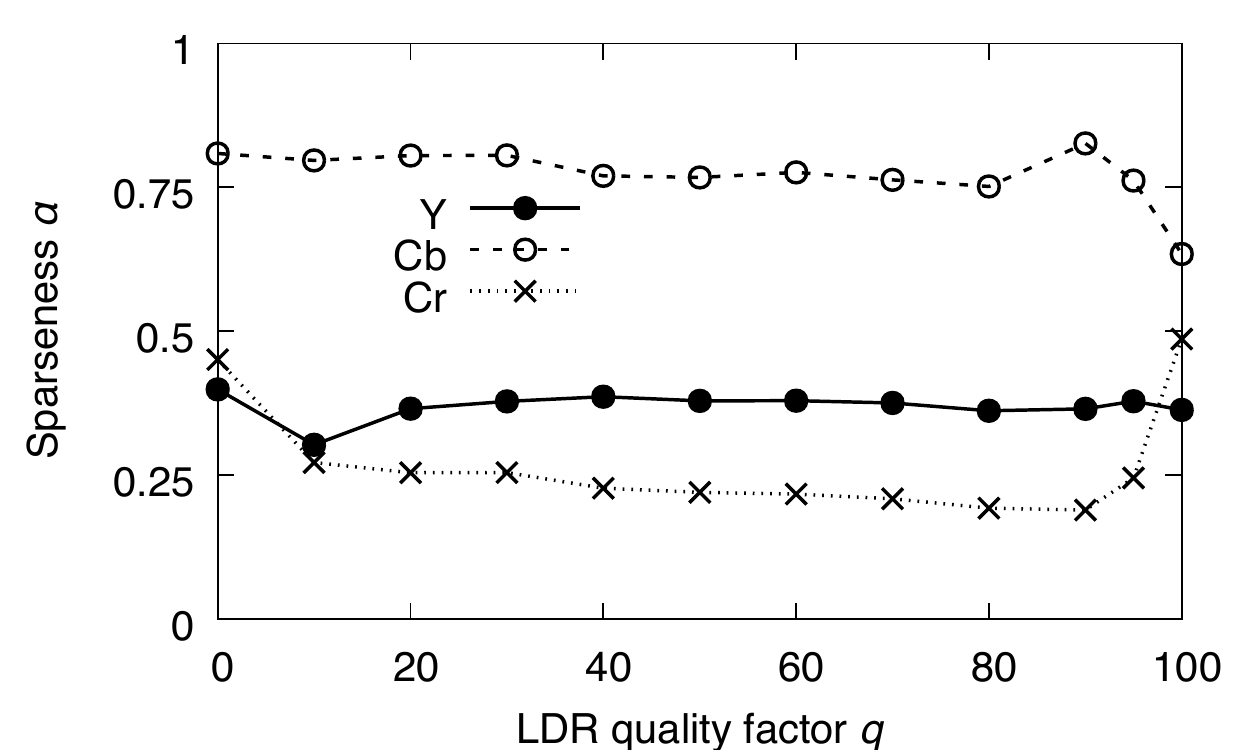}
 }
 \subfigure[MtTamWest]{
 \includegraphics[width=\histW\columnwidth]{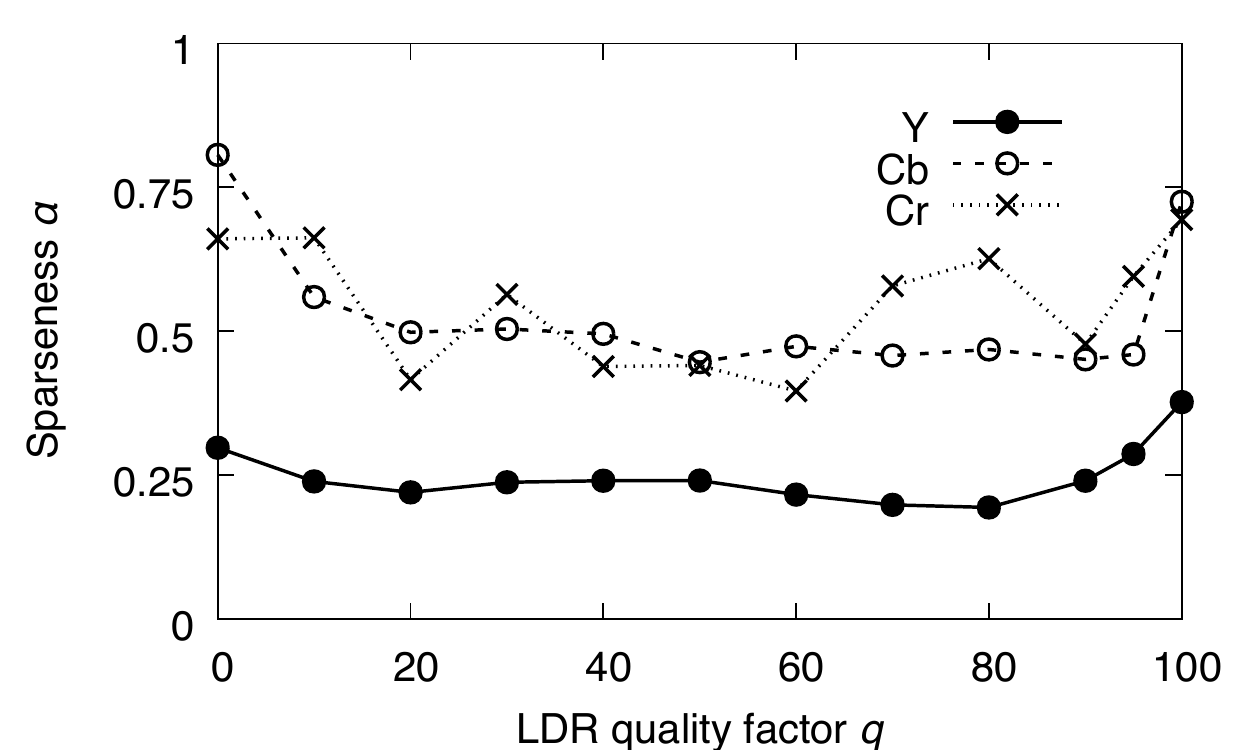}
 }
 \subfigure[Desk]{
 \includegraphics[width=\histW\columnwidth]{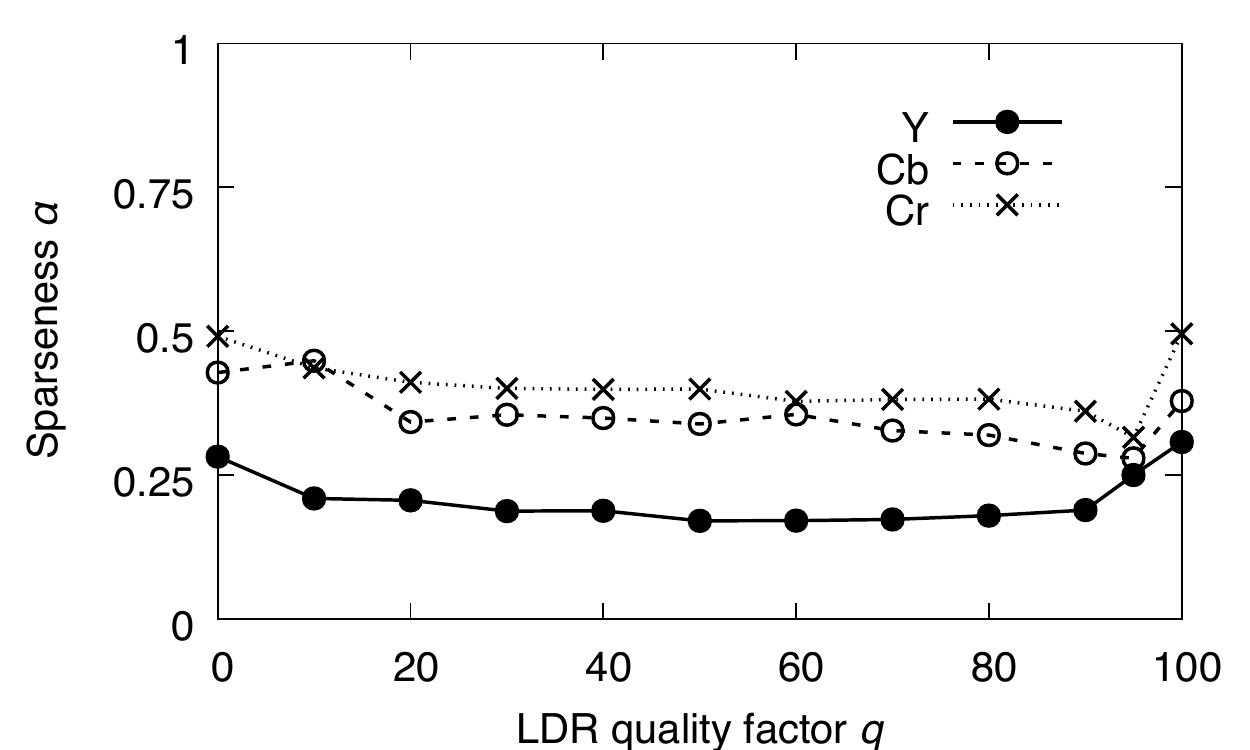}
 }
 \caption{Histogram sparseness of residual data (\myhlf{16 bit}{D14} floating-point)}
 \label{histsparse_float}
\end{figure}
\begin{figure}[tb]
 \centering
  \newcommand{\histW}{0.75}
\subfigure[Books]{
 \includegraphics[width=\histW\columnwidth]{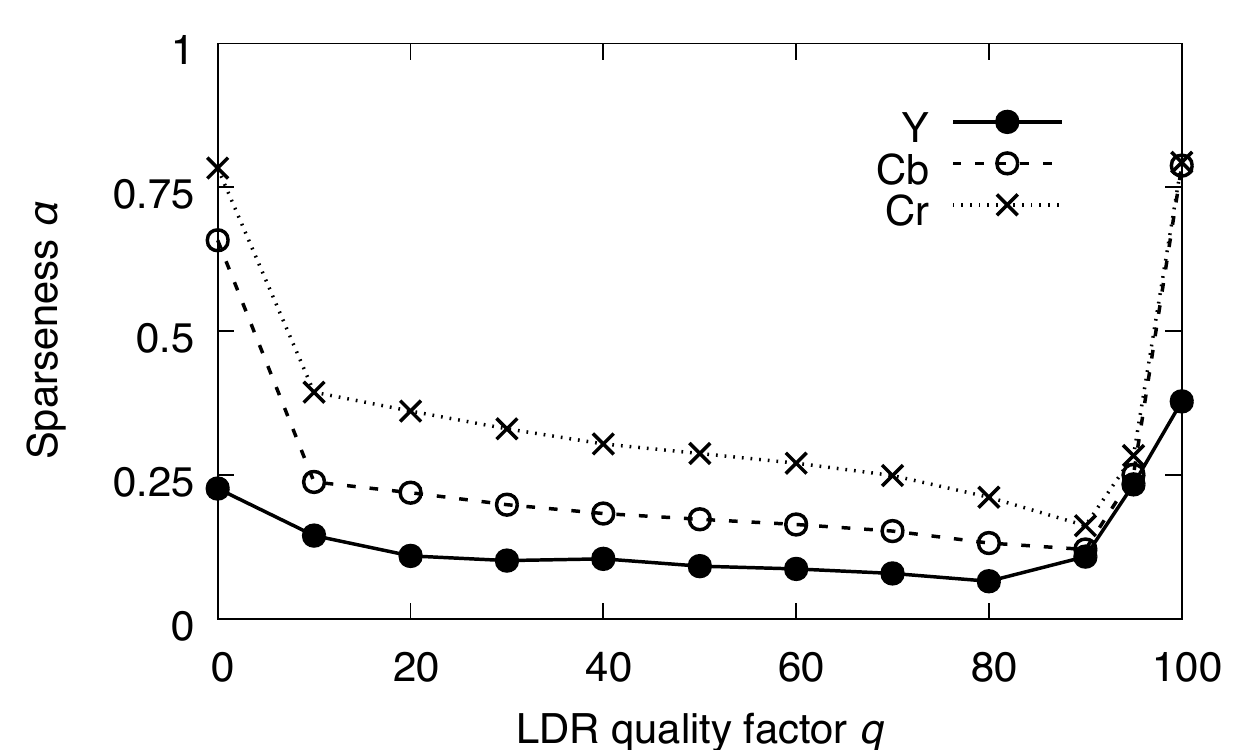}
 }
 \subfigure[Kimono]{
 \includegraphics[width=\histW\columnwidth]{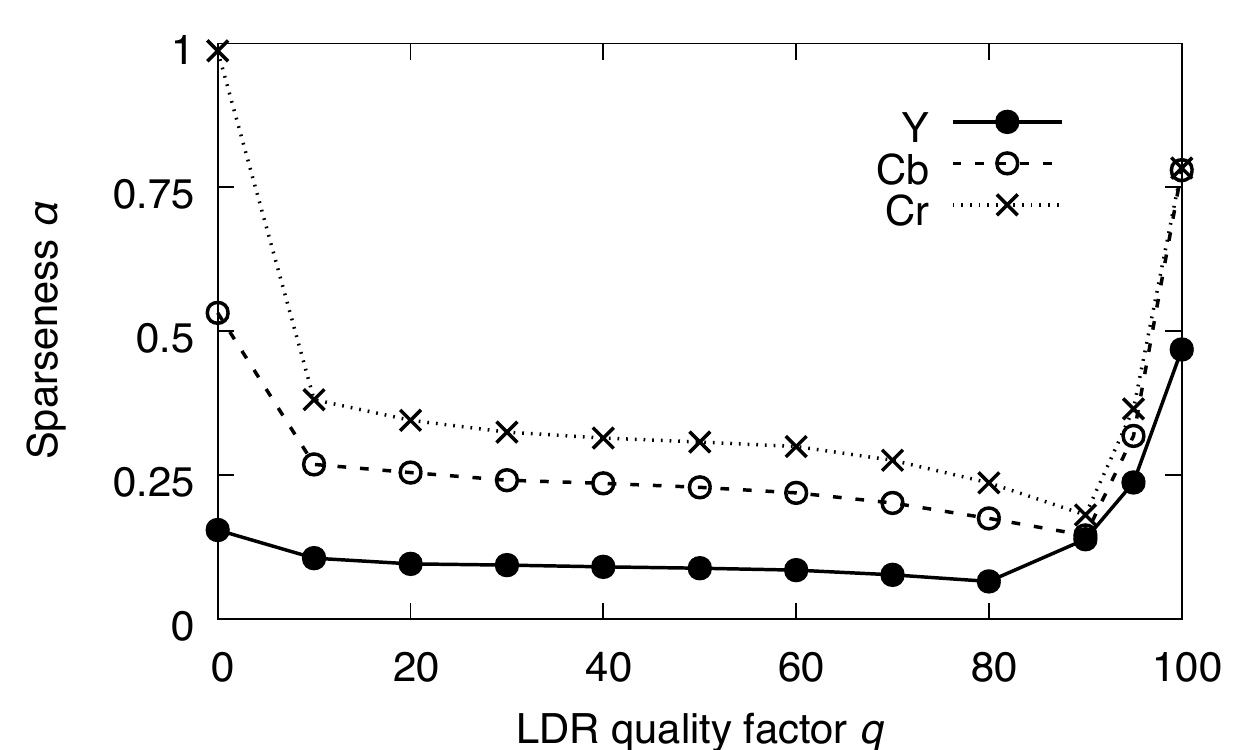}
 }
  \subfigure[Moss]{
 \includegraphics[width=\histW\columnwidth]{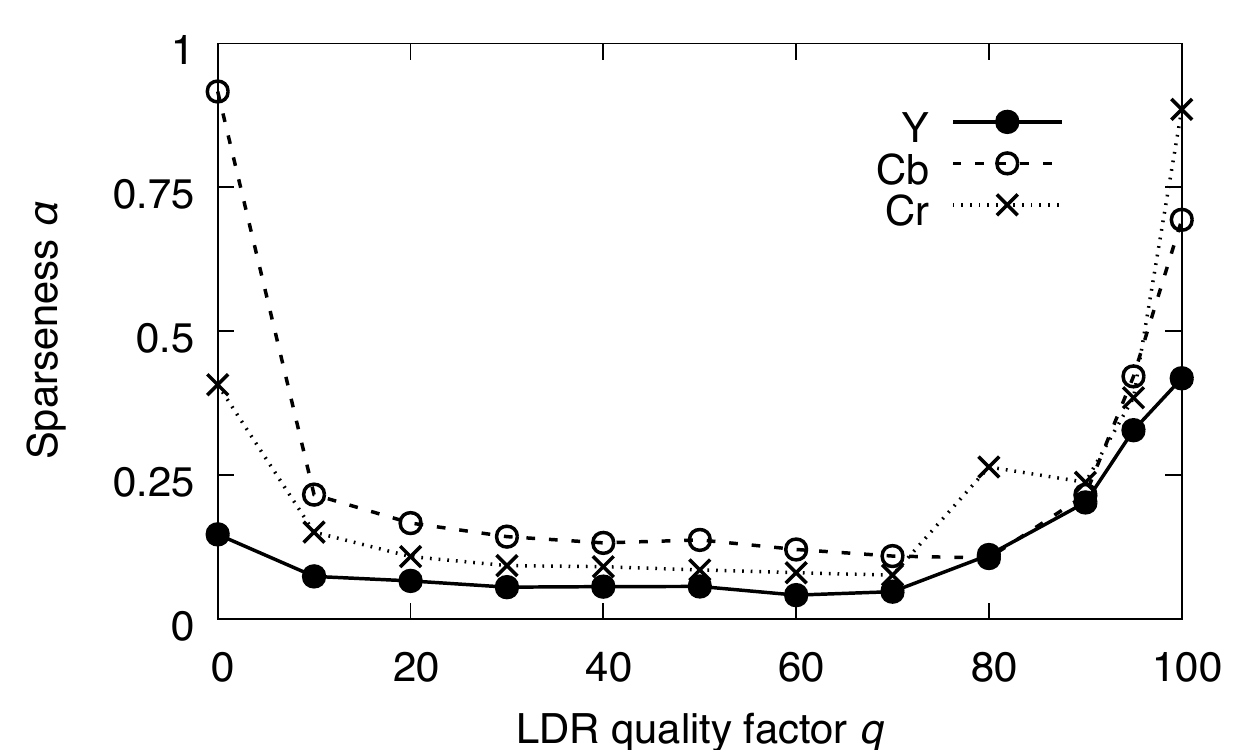}
 }
 \subfigure[MusicBox]{
 \includegraphics[width=\histW\columnwidth]{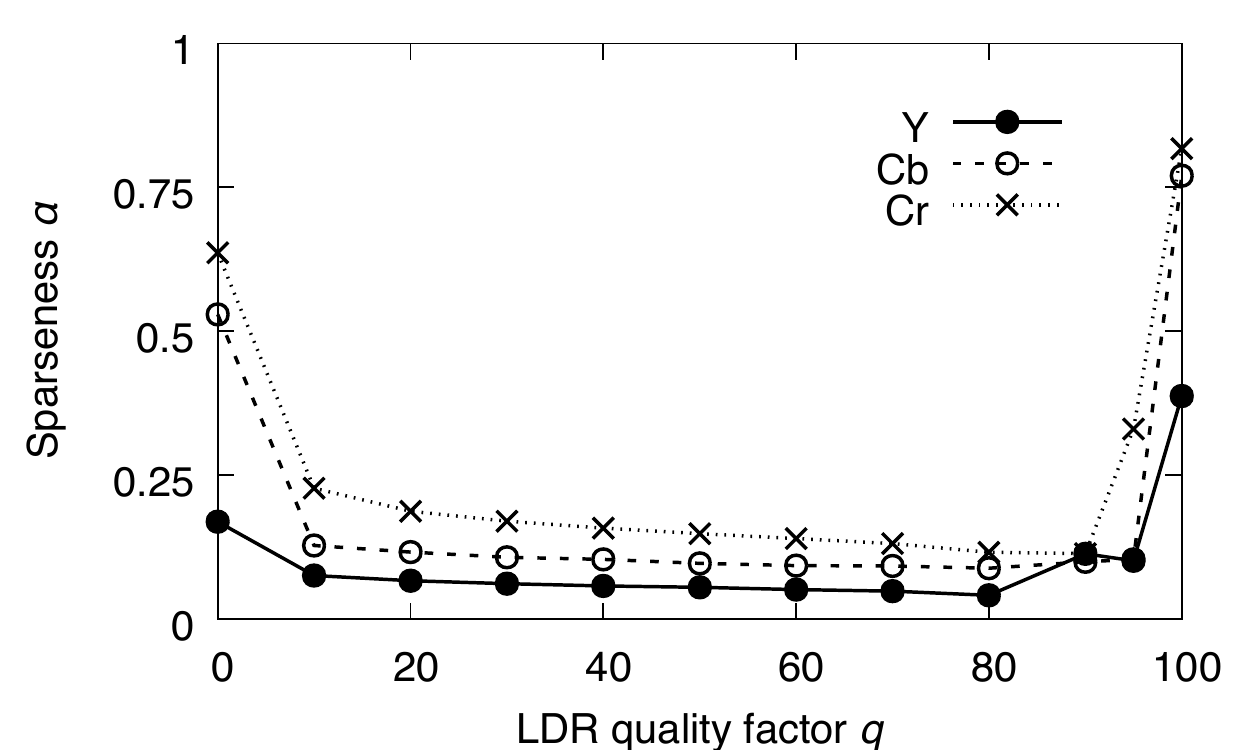}
 }

 \caption{Histogram sparseness of residual data (\myhlf{16 bit}{D14} integer)}
 \label{histsparse_int}
\end{figure}
HDR images often have sparse histograms due to its high dynamic range of
pixel values\cite{6637869}. Moreover, the histograms of the residual data in the
two-layer coding in the \myhl{Part 8}{D27} are also sparse after subtraction of
LDR data in the base layer.
In this paper,
this histogram sparseness is denoted as $\alpha$
\newsavebox{\boxe}\sbox{\boxe}{\cite{6288143}}%
 \renewcommand{\fboxsep}{0pt}
\renewcommand{\fboxrule}{0pt}
\myhl{. The histogram sparseness was originally proposed in
 Ref. {\usebox\boxe}. Let $H(x)$ denotes the frequency of a pixel value
 $x$ of an image.
 A set of pixels that have non-zero frequency is
 defined by}{A1}
 \ifannotation
 \renewcommand{\theequation}{\colorbox{yellow}{\arabic{equation}}}
 \else
 \fi
   \begin{ybox}
    X &= \{x|H(x)\neq 0\}.
   \end{ybox}

  \myhl{Then, the range of $x$ can be defined as}{A1}
    \begin{ybox}
     D(x) &= \max(x)_{x\in X} - \min(x)_{x\in X} + 1.
    \end{ybox}
  \myhl{ Let $|X|$ denotes the total number of all the elements of a set $X$.
  The histogram sparseness $\alpha$ is represented by}{A1}
   \begin{ybox}
    \alpha &= \frac{|X|}{D(x)}.
   \end{ybox}
The range of
$\alpha$ is $0\leq \alpha \leq 1$ and \myhl{the smaller}{A1} $\alpha$ means the
 sparser histogram. 
 \myhl{For example, $\alpha = 0.5$ means half of pixels within the range $D(x)$ has
 zero bin in the histogram $H(x)$ and $\alpha=1.0$ means there is no
 zero bins in the histogram $H(x)$ within the range $D(x)$.}{A1}
Figure \ref{histsparse_float} and \ref{histsparse_int} show the {\it `sparseness'} of the residual data of
two types of HDR images having floating-point and integer pixel values. The remarks from these figures are
summarized \myhl{as follows.}{D13}
\begin{itemize}
\item The sparseness depends on images and the quality factor $q$ for base layer. 
 \item The histogram of residual data tends to be sparse, especially,
      \myhl{ the luminance components ($Y$) has sparser histograms than that of
      the chroma components ($C_b$, $C_r$).}{A1,D13}
\end{itemize}

For image signals having \myhl{such sparseness}{D13}, it is well known that the
histogram packing technique improves lossless
compression
performance\cite{958146,1034993,988715,1040040,ELCVIA116,eusipco2017}.
\newsavebox{\boxf}\sbox{\boxf}{\cite{958146,1034993,988715,1040040,ELCVIA116,eusipco2017}}%
\myhl{
In Refs. {\usebox\boxf}, the sparseness $\alpha$ was increased and the
range $D(x)$ became narrower by using
histogram packing. It was also reported that the lossless image compression performance
improved for histogram packed images.}{A1}
The main idea of the proposed
method is to combine the two-layer coding structure with the histogram
packing technique.
\subsection{Histogram packing}

\begin{figure}[tb]
 \centering
 \ifannotation\colorbox{yellow}{
 \else
 \fi
 \includegraphics[width=1.0\linewidth]{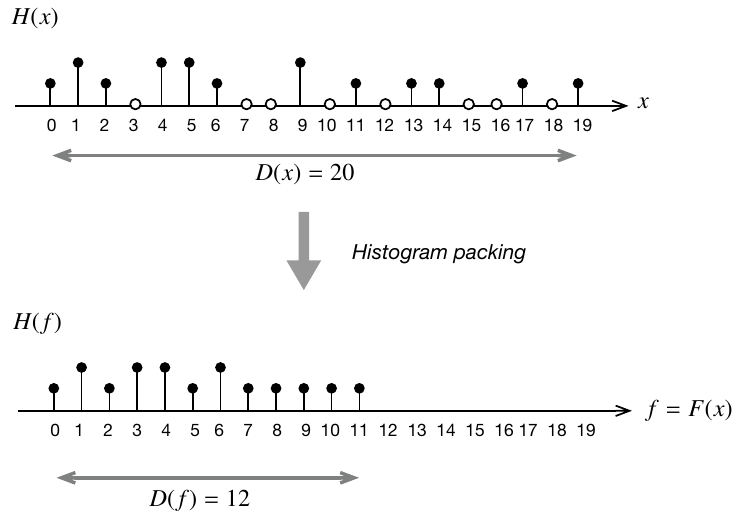}
 \ifannotation
 }
 \else
 \fi
 \caption{\myhlf{Procedure of histogram packing}{A2-2}}
 \label{howtohistopack}
\end{figure}

   \begin{figure}[tb]
\centering
    \ignore{
   \centering
   \subfigure[Entire range (Sparseness $\alpha=0.374$)\label{entire}]{
   \includegraphics[width=1.0\linewidth]{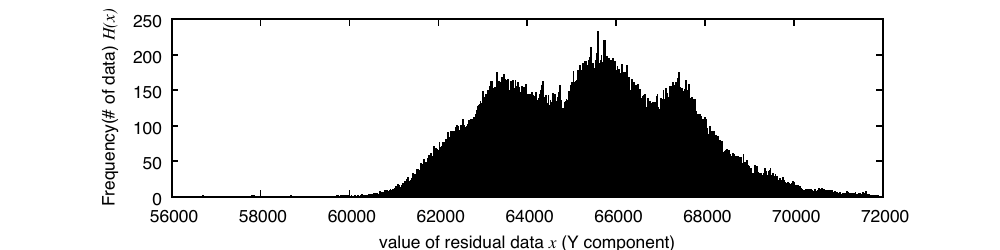}
   }
   \subfigure[ $65000\leq x \leq 66000$ (Sparseness $\alpha=0.5$)\label{65000}]{
   \includegraphics[width=1.0\linewidth]{./hkob/fig3fig4/h65000.pdf}
   }
   \subfigure[$60000\leq x \leq 61000$ (Sparseness $\alpha=0.048$)\label{60000}]{
   \includegraphics[width=1.0\linewidth]{./hkob/fig3fig4/h60000.pdf}
   }
   \subfigure[$57000\leq x \leq 58000$ (Sparseness $\alpha=0.354$)\label{57000}]{
   \includegraphics[width=1.0\linewidth]{./hkob/fig3fig4/h57000.pdf}
    }
    }
    \includegraphics[width=1.0\linewidth]{./hkob/fig3fig4/hall.pdf}
   \caption{Histograms for residual data (Y component of \myhlf{`BloomingGorse2'}{D28},
    LDR $q=50$)}
   \label{partialHisto}
   \end{figure}
  
  \begin{figure}[tb]
   \subfigure[Index image (after histogram packing)]{
   \includegraphics[width=0.47\linewidth]{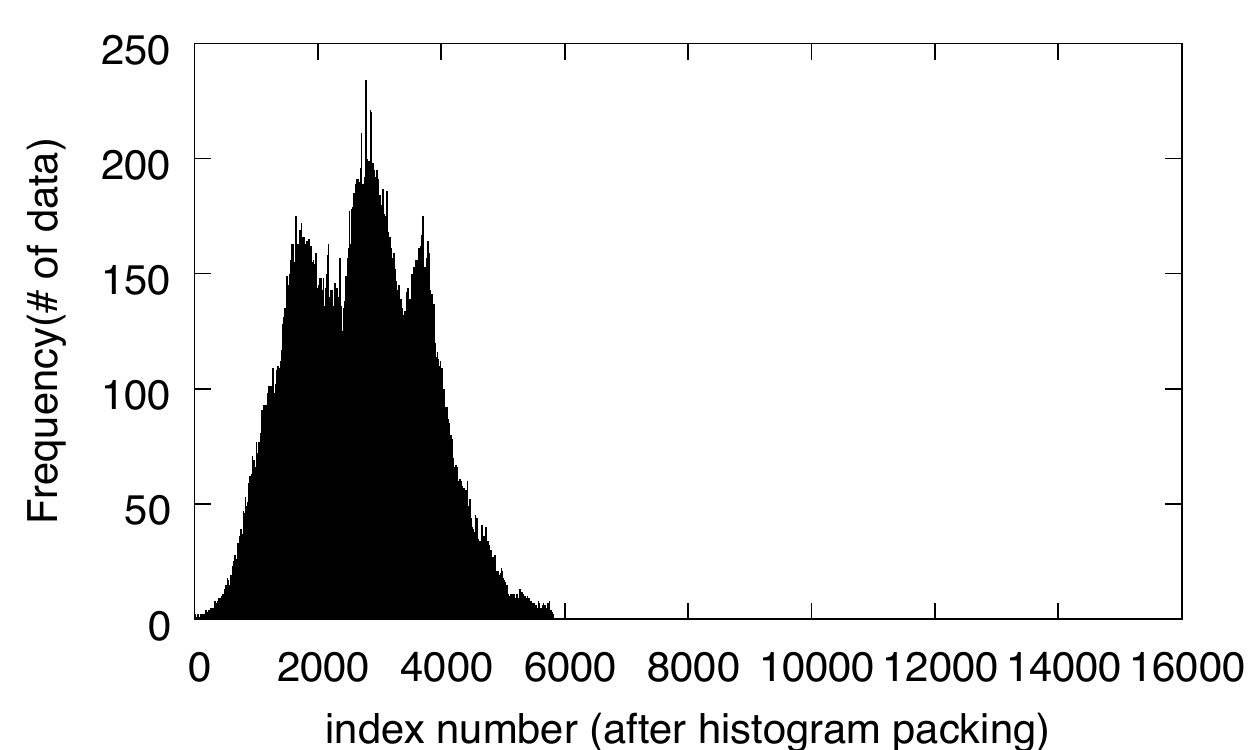}
   \label{idximg}
   }
   \subfigure[Unpacking table]{
   \includegraphics[width=0.47\linewidth]{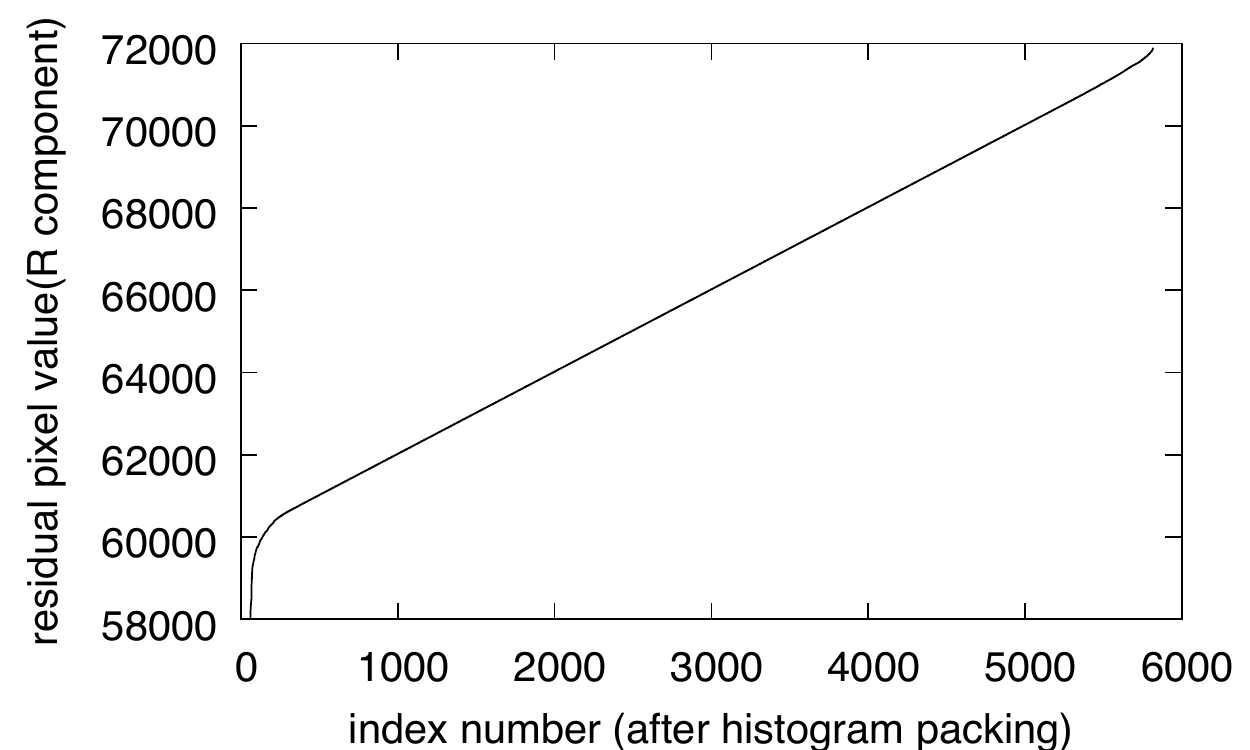}
   \label{utable}
   }
  \caption{Histogram of index image and unpacking table (Y
   component of \myhlf{`BloomingGorse2'}{D28}, LDR $q=50$)}
   \label{packedIdxTbl}
  \end{figure}
\myhl{In section 3.1}{D15}, it has been noted that the histograms of the residual data tend to be
sparse and the \myhl{increase}{A2-1} of the sparseness is effective to improve
lossless coding performance%
\newsavebox{\boxg}\sbox{\boxg}{\cite{958146,1034993,988715,1040040,ELCVIA116,eusipco2017}}%
\myhl{{\usebox\boxg}.}{A2-1}
\newsavebox{\boxj}\sbox{\boxj}{\cite{6288143}}%
\myhl{
The histogram packing {\usebox\boxj} maps a pixel value $x$ to $f$
according to
}{A2-2}
 \begin{ybox}
  f&=F(x),
 \end{ybox}
 \myhl{%
 where}{A2-2}%
  \begin{ybox}
   F(x)=
\begin{cases}
0 & \text{for $x =$ min,}\\
F(x-1) & \text{for $x > $ min $\wedge H(x) = 0$,} \\
F(x-1)+1 & \text{for $x >$ min $\wedge H(x) \neq 0$.} 
\end{cases}
  \end{ybox}
 \newsavebox{\boxk}\sbox{\boxk}{\ref{howtohistopack}}%
\myhl{
As illustrated in Fig. {\usebox\boxk}, histogram packing converts
$x$ into $f$, and the histogram sparseness is used to reduce the range
from $D(x)=20$ to $D(f)=12$. This operation is
 reversible{\usebox\boxj}.}{A2-2}
 
 
Figure \ref{partialHisto} shows histograms
$H(x)$ for Y component in the residual data of an HDR image
\myhl{`BloomingGorse2'.}{D28} Horizontal axes denote the pixel values $x$ in
re-interpreted integer
number of floating-point representation described \myhl{in section 2.}{D15}
%
After histogram
packing, a histogram-packed image is obtained. In this paper, this
histogram-packed image is
referred to as \myhl{`index image'}{D28}. \myhl{$H(f)$}{A2-2} for the
index image is shown in Fig. \ref{idximg}\myhl{. It is clearly considered
to be dense and the sparseness $\alpha=1.0$.}{A2-2}
\myhl{The index image is compressed by an arbitrary lossless image
encoder, e.g. JPEG 2000 or JPEG XR etc.}{A2-2}
The unpacking table, which is necessary to perform inverse histogram
packing, is illustrated in Fig. \ref{utable}. 
\myhl{Obviously, it is considered as one-to-one correspondence
function. This table is effectively compressed by using DPCM because it
is monotonically increasing.}{B3}

  \subsection{Encoder structure}
\begin{figure*}[tb]
 \centering
 \ifannotation
 \includegraphics[width=1.35\columnwidth]{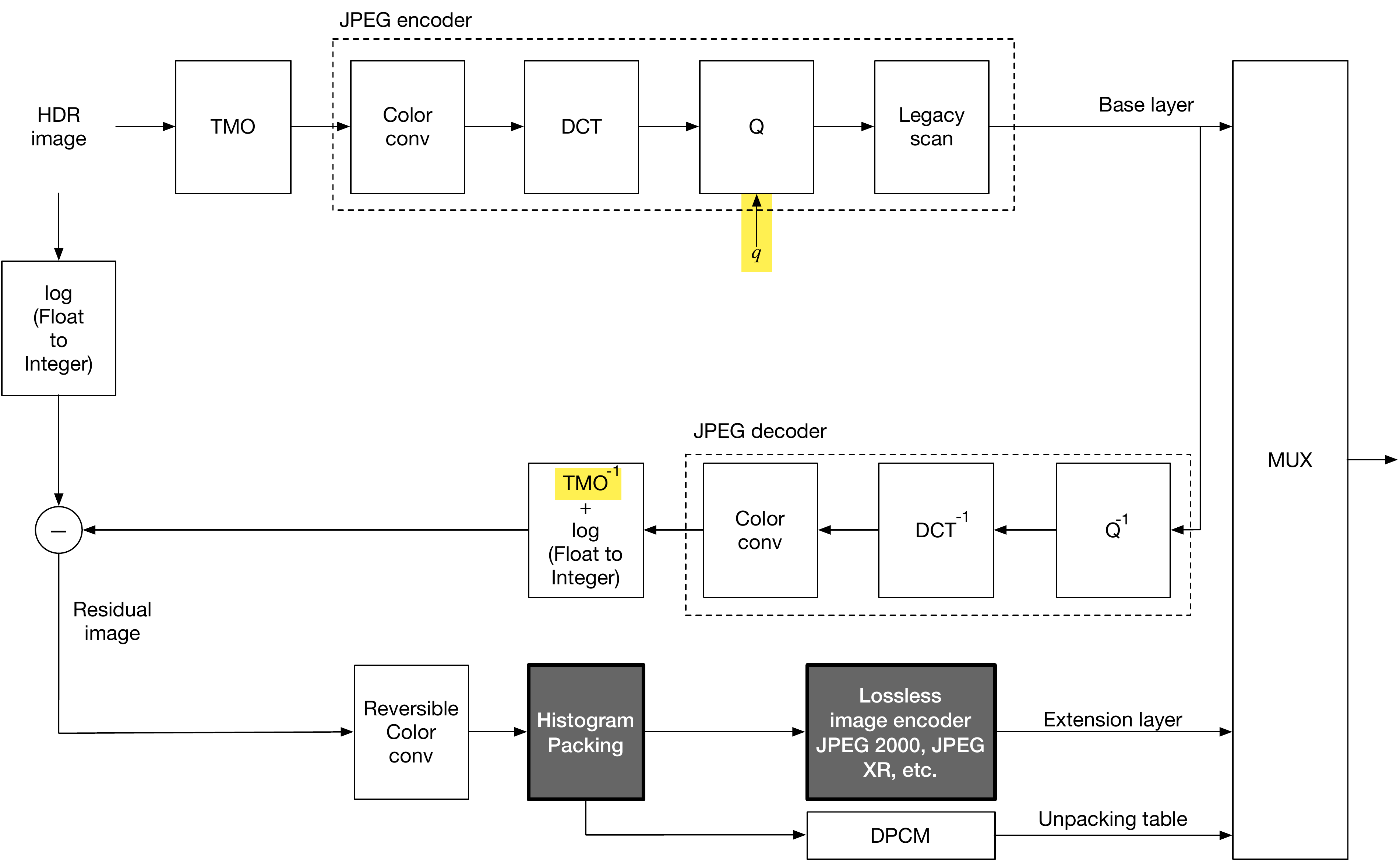}
 \else
 \includegraphics[width=1.35\columnwidth]{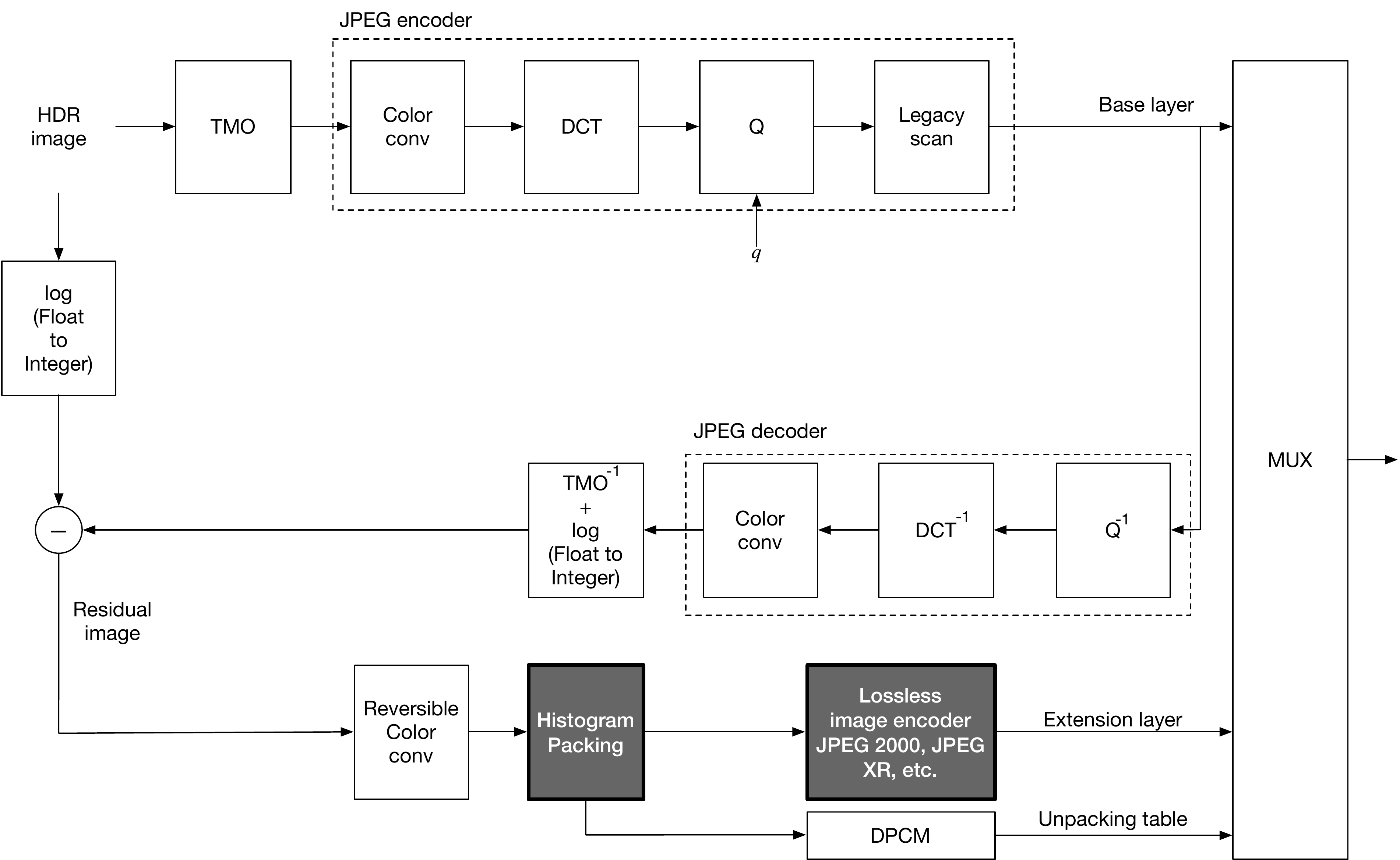}
 \fi
 \caption{Blockdiagram of proposed lossless encoder}
 \label{proposed_enc}
\end{figure*}
The structure of the proposed lossless two-layer coding is illustrated in
Fig.\ref{proposed_enc}. The coding-path to generate a base layer, which
is backward compatible with the legacy JPEG, is the exactly same as the
\myhl{JPEG XT Part 8}{D27}
\myhl{except that}{C1,D16} the refinement scan for the base layer is not
used. Therefore, \myhl{the LSB split-off in the JPEG XT Part 8 is not used and}{C1,D27} the value of $R$ is set to zero.

For the extension layer, which consists of the residual data generated
by subtracting partially decoded base layer from the original HDR image, the
coding procedure after color space conversion from RGB to YCbCr is
different from the \myhlf{Part 8}{D27} encoder %
\newsavebox{\boxa}\sbox{\boxa}{\ref{proposed_enc}}%
\myhlf{as depicted in Fig. {\usebox\boxa}.}{D17}
The histogram of each
color component of the color converted residual data is analyzed and packed by
using the histogram packing technique. Then, the packed residual data is
compressed
\myhl{by an arbitrary lossless image encoder.}{D18}
After the
subtraction described above, the residual data for each color component
may have 17 ($=16+1$) bit integers. This over 16 bit in the bit-depth is the reason
for using such lossless encoders as the JPEG 2000 and the JPEG XR \myhl{in
this paper}{D19}
because they are able to accept up to 32 bit integer pixel value
per component\cite{1528434}.

For the inverse operation of the histogram packing, unpacking table is
sent to the decoder. The unpacking table is one-to-one correspondence
function between the packed index value and the original pixel
value. Since this is monotonically increasing, DPCM is performed an then
\myhl{the unpacking table encoded by using DPCM}{D29} is compressed by bzip2\cite{bzip2} algorithm
to reduce the amount of data.

Finally, the base layer which is compatible with the legacy JPEG, the extension
layer consists of the lossless JPEG 2000 or JPEG XR codestream, and
\myhl{unpacking table compressed by using bzip2 algorithm}{D29} are multiplexed into a codestream and \myhl{it shall be sent to}{D20} the
decoder. Note that the proposed method does not need to adjust the 
value of the coding parameters to meet the input image and to get sufficient lossless
coding efficiency. The LDR quality $q$ is the only parameter to be
determined according to the user's demand. Thus, the
proposed method can be considered to be image-independent and almost-parameter-free.

\section{Experimental results}
To verify the effectiveness of the proposed method, the lossless compression performance in terms of
bitrate of the generated codesrtreams was evaluated and compared with
that of the \myhl{JPEG XT Part 8}{D27}.
\subsection{Conditions}
\subsubsection{Test images}
 \begin{table}[tb]
  \centering
  \caption{Test images (bpp means bit-depth per component): All images have three color components in RGB color space.}
\label{testimages}
   \begin{tabular}{c|c|c|c|c}
    \hline
   Pixel value type & Index & Name & bpp & Size \\\hline
  \multirow{4}{*}{Floating-point}& f1 & memorial & 16 & 512$\times$768\\
  & f2& Blooming Gorse2& 16 & 4288$\times$2848\\
  & f3& MtTamWest & 16 & 1214$\times$732\\
  & f4& Desk& 16 &644$\times$874\\ \hline
  \multirow{4}{*}{Integer}& i1 & Books & 12 & 3840$\times$2160\\
  & i2& Kimono & 12 & 3840$\times$2160\\
  & i3& Moss & 12 &3840$\times$2160\\
  & i4& MusicBox& 12 & 3840$\times$2160\\ \hline
   \end{tabular}
 \end{table}
Images having both floating-point and integer pixel values were selected for
the experiments. For floating-point images, four of images common to HDR related
experiments were collected.
For integer images, four of the ITE test
images\cite{ITEIMG} were selected. The specifications of these test images are summarized in
Table \ref{testimages}. 
Although some of floating-point images have full precision float
\myhl{value for each pixel}{D21}
, we have converted the values into half precision float because
the JPEG XT encoder only accepts half precision floating-point pixels as
its inputs.
Note that image names are all represented by the index shown in
Table\ref{testimages}. The first character of the index means the type
of pixel values;
\myhl{``f'' and ``i'' stand for floating-point and integer respectively,
the following number of the first character corresponds to each image name.}{D22}
\subsubsection{Encoder software}
For the \myhl{JPEG XT Part 8}{D27} encoder, the reference software\cite{JPEGXTpart5,JPEGXTSOFT} available from
the JPEG committee was used. For the proposed method, the modified
encoder of the reference software, whose coding path for the residual
data was changed to have the histogram packing and JPEG 2000/JPEG XR encoder,
was used. The Kakadu software\cite{Kakadu} and the
reference software of the JPEG XR\cite{JPEGXRSOFT} were used as those encoders that were used to compress the
histogram-packed residual data.
The lossless performances of the proposed method and the \myhl{JPEG XT Part 8}{D27} were evaluated with several values of $q$ (quality factor of LDR
image) and $R$ (number of refinement bits for base layer). For the JPEG
XT, another parameter, the effect of $rR$ (number of refinement bits for extension
layer), was also evaluated.
\subsection{Results and remarks}
\subsubsection{Overall lossless performance}
 \begin{figure}[tb]
  \centering
  \subfigure[Floating-point\label{overall_float}]{
  \includegraphics[width=0.8\linewidth]{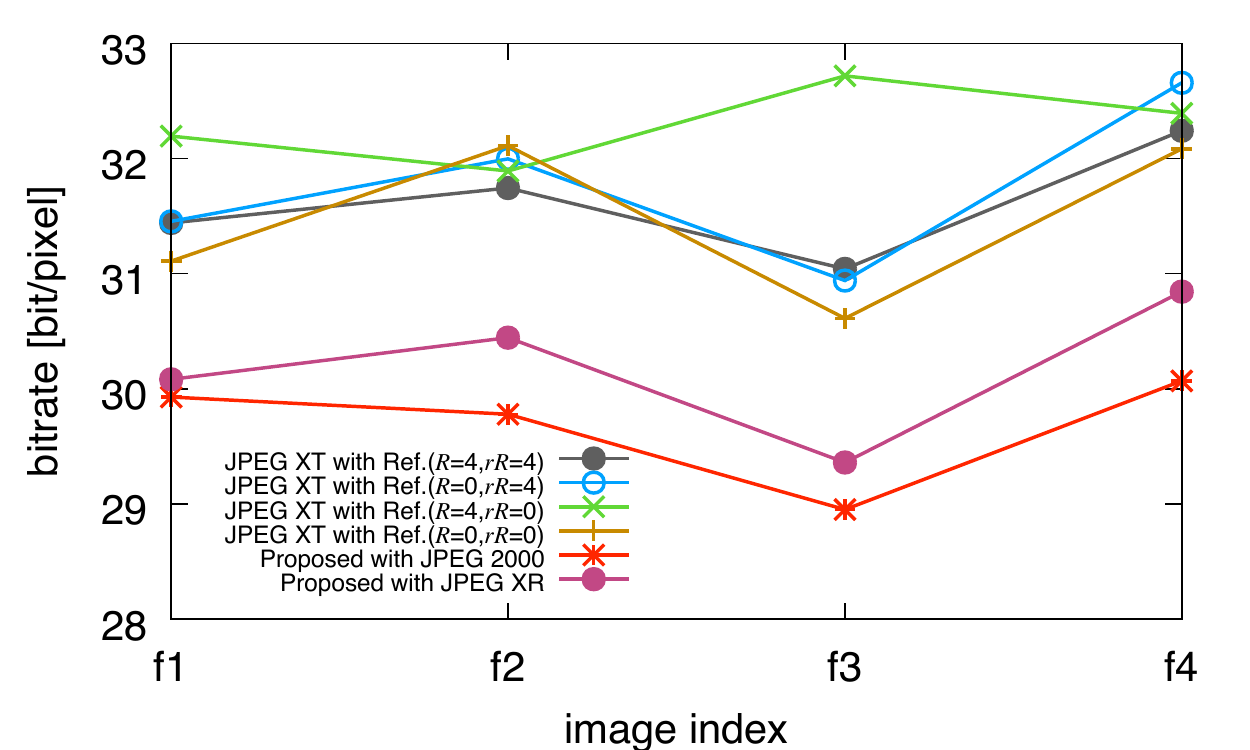}
  }
  \subfigure[Integer\label{overall_int}]{
  \includegraphics[width=0.8\linewidth]{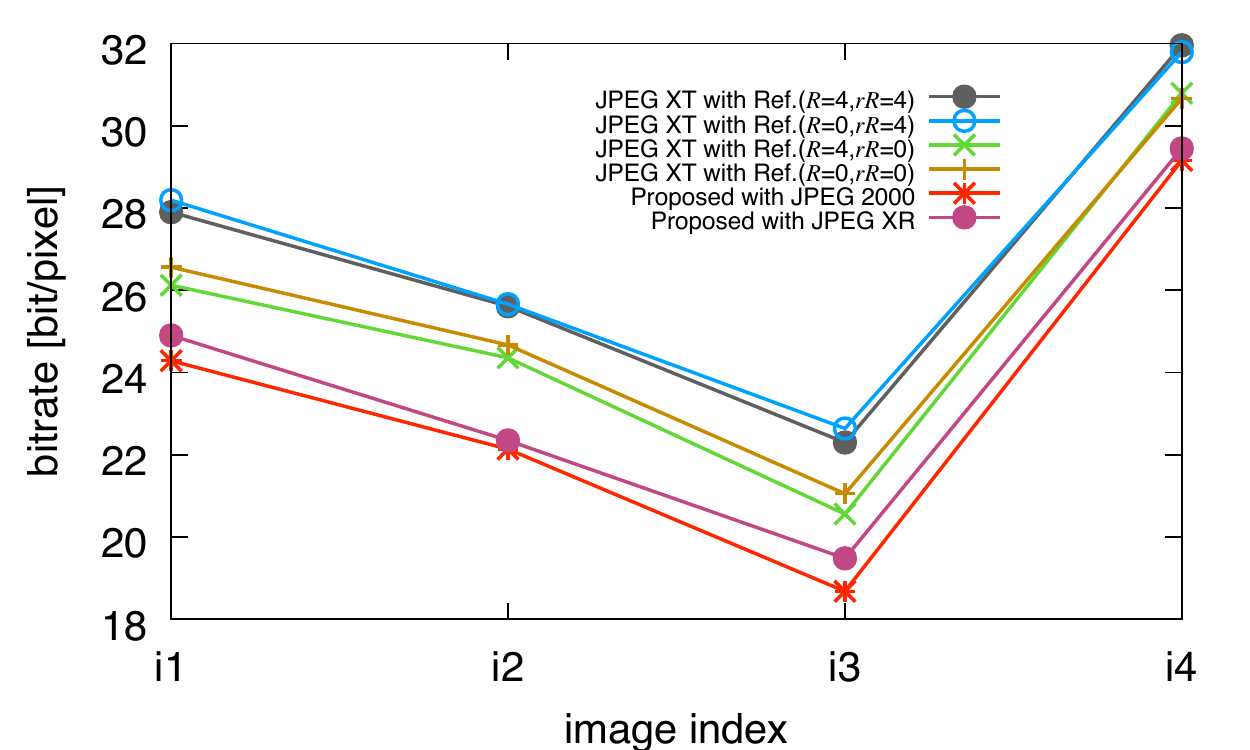}
  }
  \caption{Comparison lossless performance between proposed method and
  JPEG XT with $q=80$: image names are represented by index (see Table. \ref{testimages}.)}
  \label{comp_xt_pro}
 \end{figure}
 Figure \ref{comp_xt_pro} shows the bitrates of lossless compressed images
 by the proposed method and the \myhl{JPEG XT Part 8}{D27} with fixed LDR quality
 $q=80$.
  \myhl{The bitrate for the proposed method includes the amount of
  unpacking table which is compressed by bzip2 algorithm.}{D23,D29}
 For the JPEG XT, the combinations of the parameters for the number of
 the refinement bits for both the base and extension layer,
 $(R, rR) =(0,0), (0,4), (4,0), (4,4)$ were used.
 \newsavebox{\boxc}\sbox{\boxc}{\ref{overall_float}}%
  \newsavebox{\boxd}\sbox{\boxd}{\ref{overall_int}}%
\myhl{Figure {\usebox\boxc} and {\usebox\boxd}}{D24}
 show the bitrate of
 lossless compressed images having floating-point and integer pixel
 values, respectively.
Among all test images, \myhl{it was confirmed that}{D25} the lossless bitrates
provided with \myhl{the proposed method were
smaller than}{D25} those with the JPEG XT.

\ignore{
   \subsubsection{Effect of refinement bits $R$ and $rR$}
 \begin{figure}[tb]
  \centering
  \subfigure[Floating-point]{
  \includegraphics[width=0.465\linewidth]{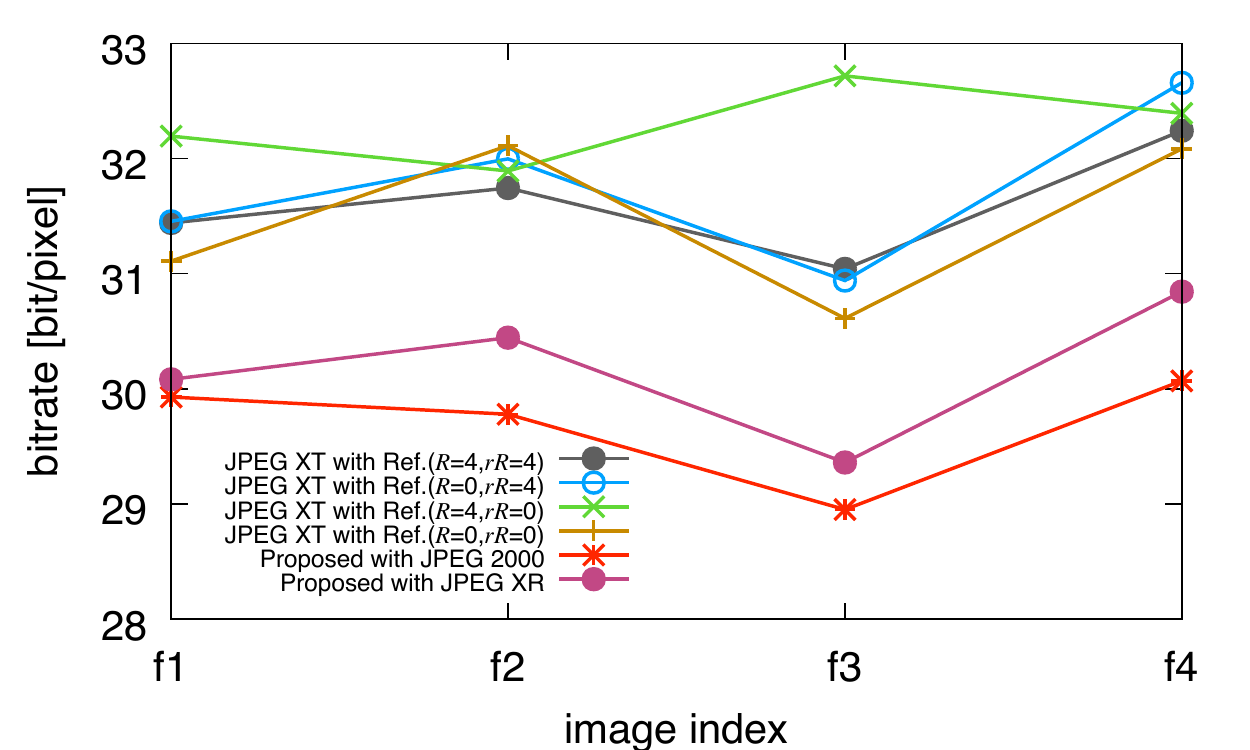}
  }
  \subfigure[Integer]{
  \includegraphics[width=0.465\linewidth]{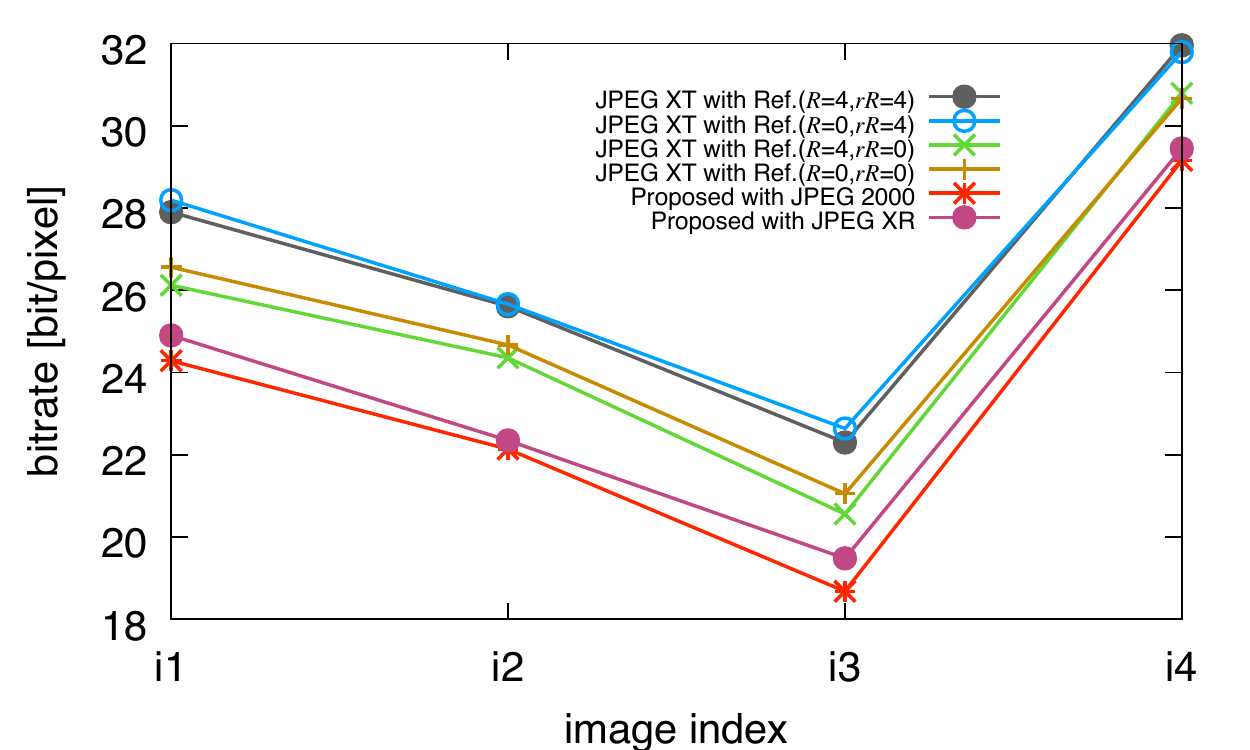}
  }
  \caption{Comparison lossless performance effect of parameter value $R$
  and $rR$
  with $q=80$: image names are represented by index (see Table. \ref{testimages}.)}
  \label{independent_xt_pro}
 \end{figure}

Figure \ref{independent_xt_pro} shows the results for the different
values of the refinement bits for base layer $R$ and the refinement bits
for the extension layer $rR$. The quality of LDR images were fixed to $q=80$.

 Moreover, to achieve good performance with
the JPEG XT

From these figures, it is verified that the proposed method shows better
lossless performance than the \myhl{JPEG XT Part 8}{D27} with any combination of the parameters.
In addition to the better performance, it is not required for the
proposed method to find the
image-dependent combination of the coding parameter
values, such as $q$, $R$ and $rR$.
}
\subsubsection{Effect of LDR quality $q$}
 \begin{figure*}[tb]
  \centering
  \subfigure[f1\label{res_f1}]{
  \includegraphics[width=0.40\linewidth]{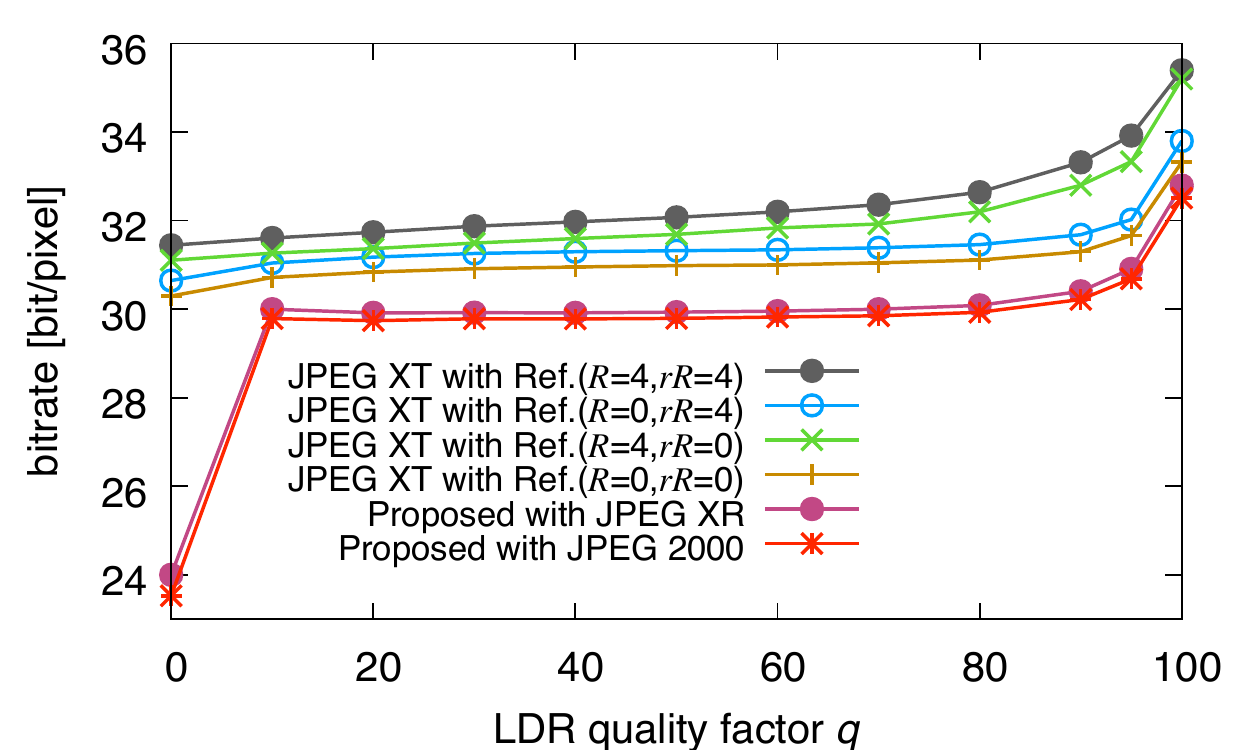}
  }
  \subfigure[f2\label{res_f2}]{
  \includegraphics[width=0.40\linewidth]{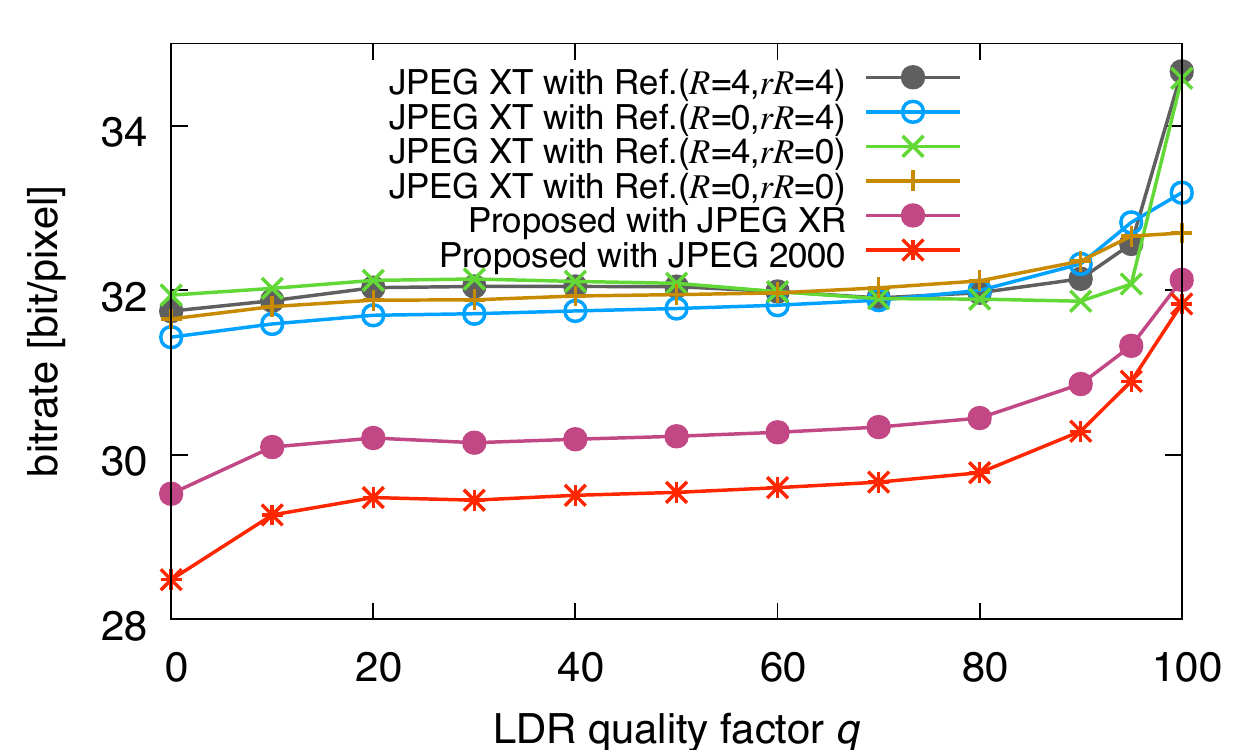}
  }
  \subfigure[f3\label{res_f3}]{
  \includegraphics[width=0.40\linewidth]{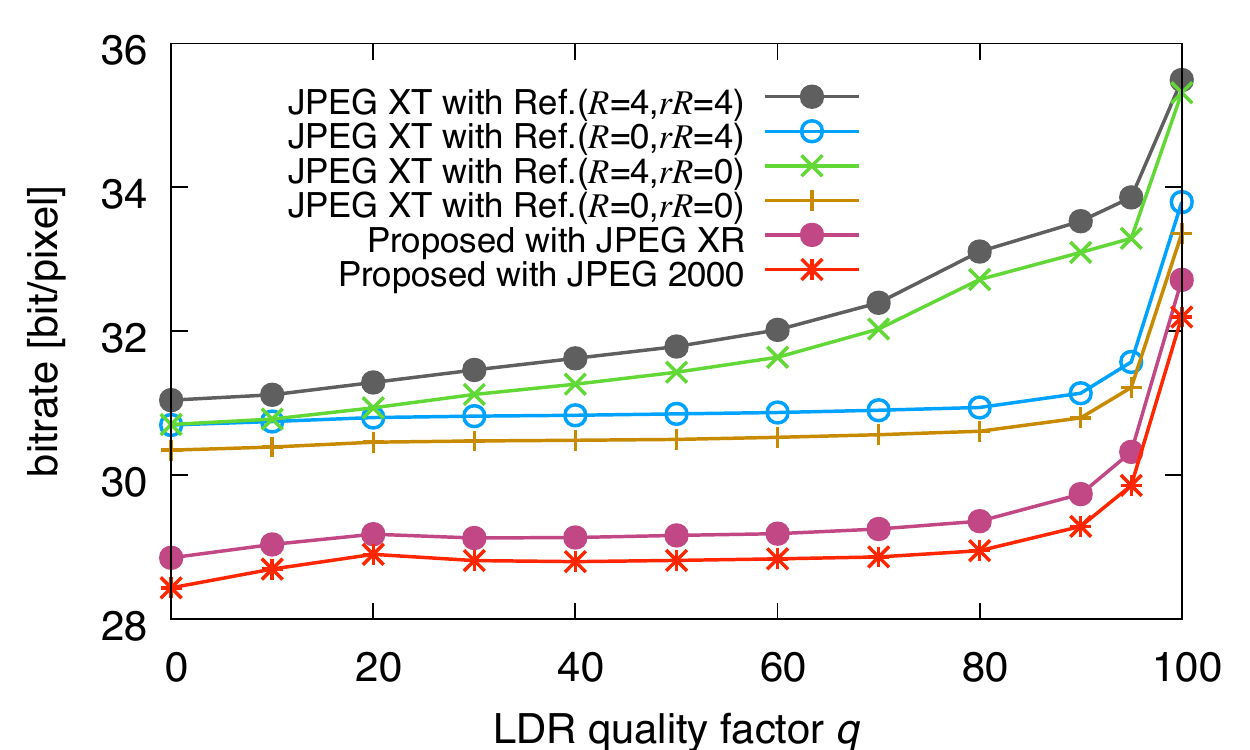}
  }
  \subfigure[f4\label{res_f4}]{
  \includegraphics[width=0.40\linewidth]{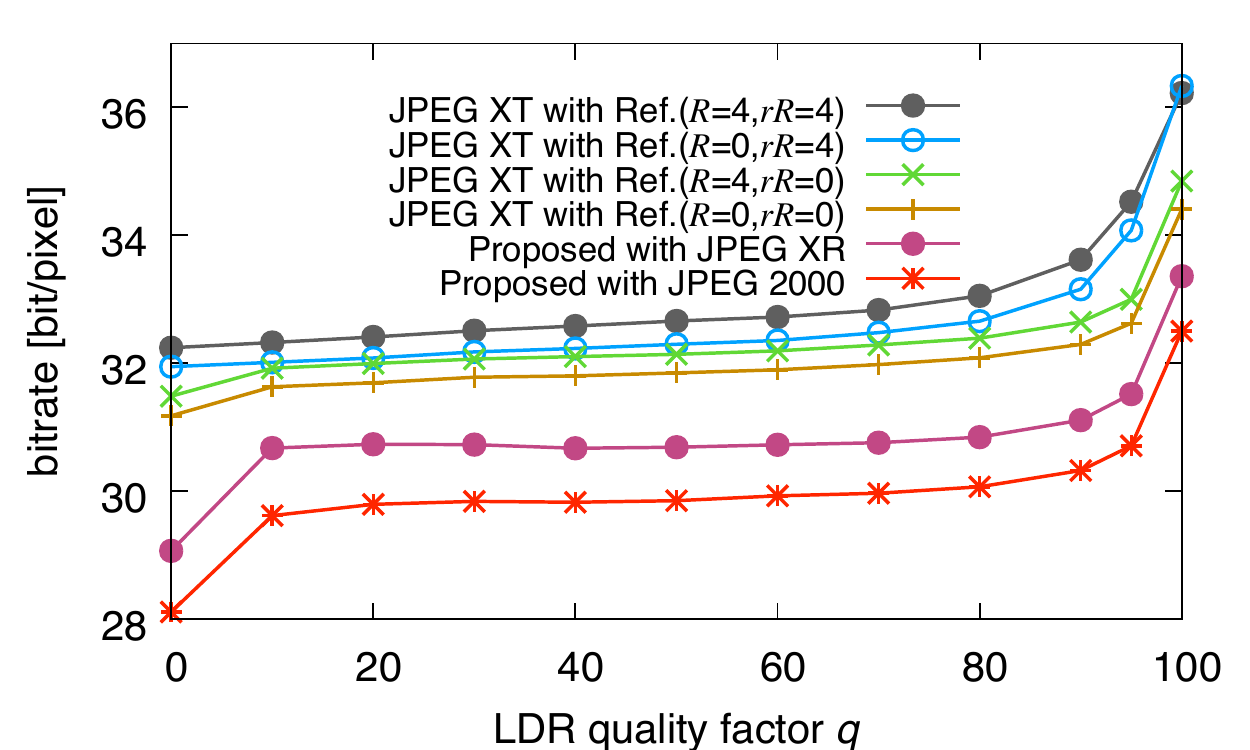}
  }
  \caption{Bitrates of lossless compressed image (float): image names are represented by index (see Table. \ref{testimages}.)}
  \label{floating_results}
 \end{figure*}
  \begin{figure*}[tb]
  \centering
  \subfigure[i1\label{res_i1}]{
  \includegraphics[width=0.40\linewidth]{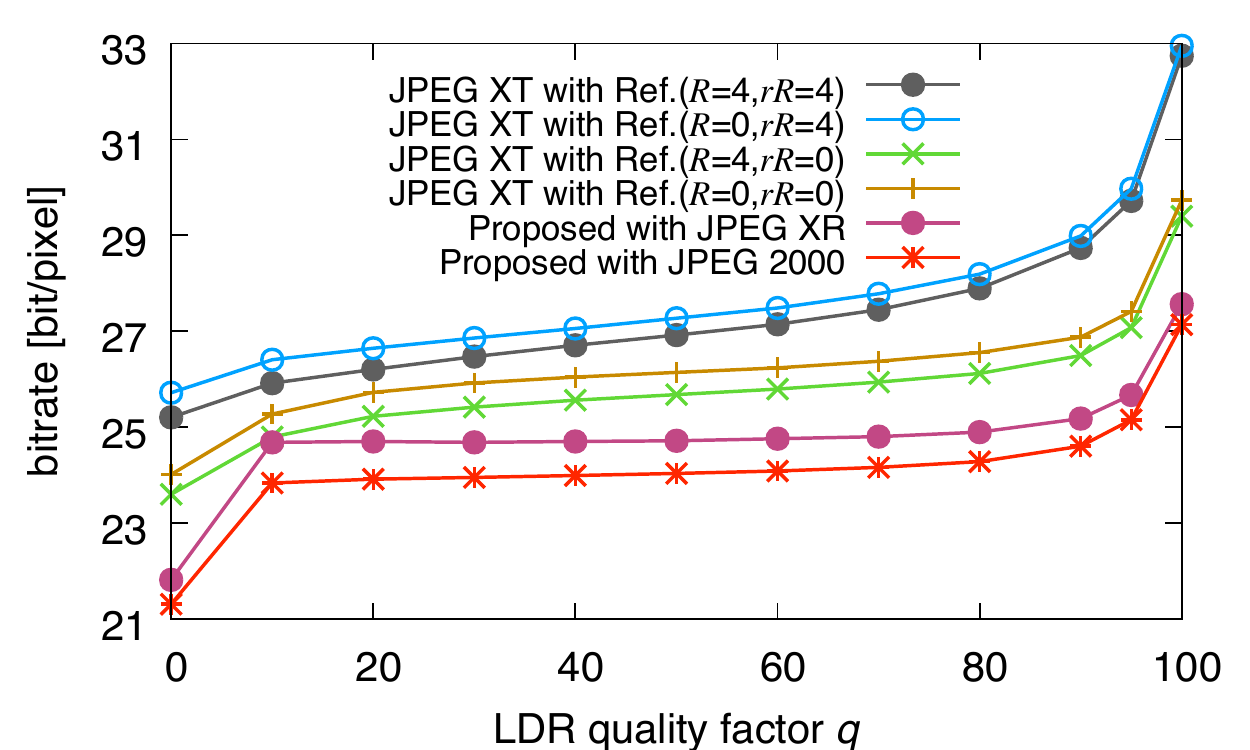}
  }
  \subfigure[i2\label{res_i2}]{
  \includegraphics[width=0.40\linewidth]{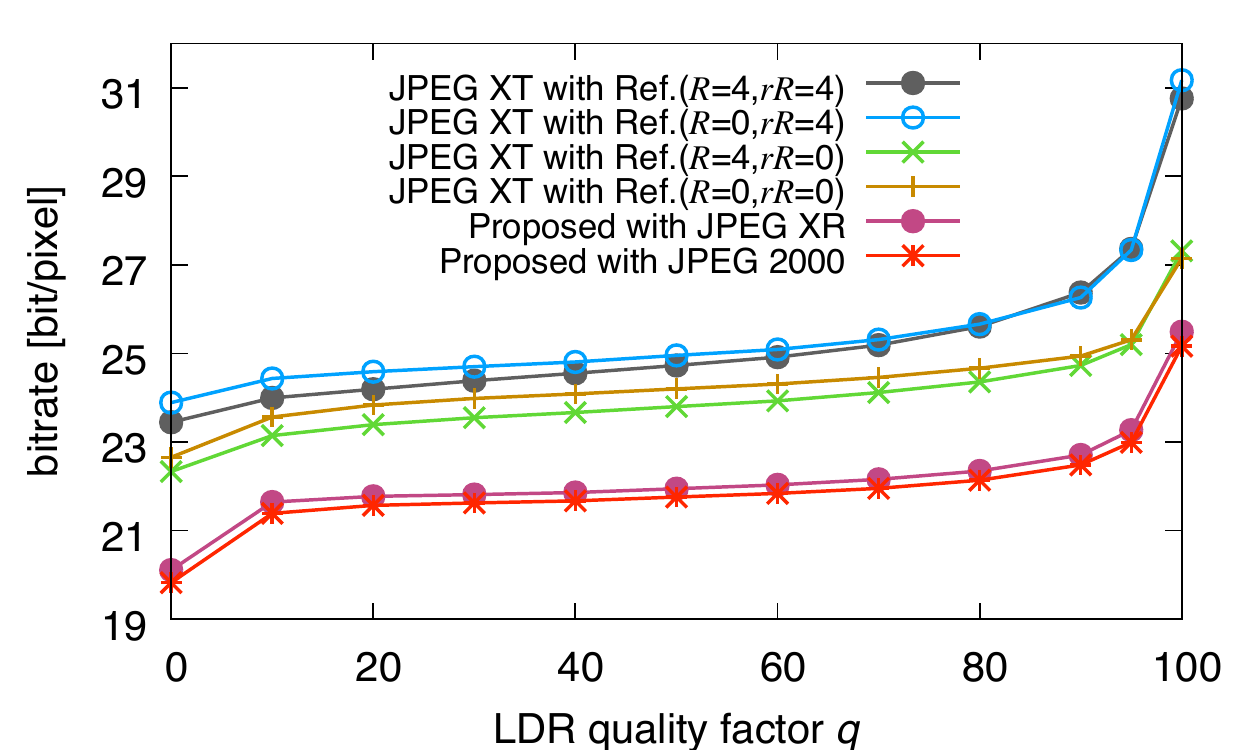}
  }
  \subfigure[i3\label{res_i3}]{
  \includegraphics[width=0.40\linewidth]{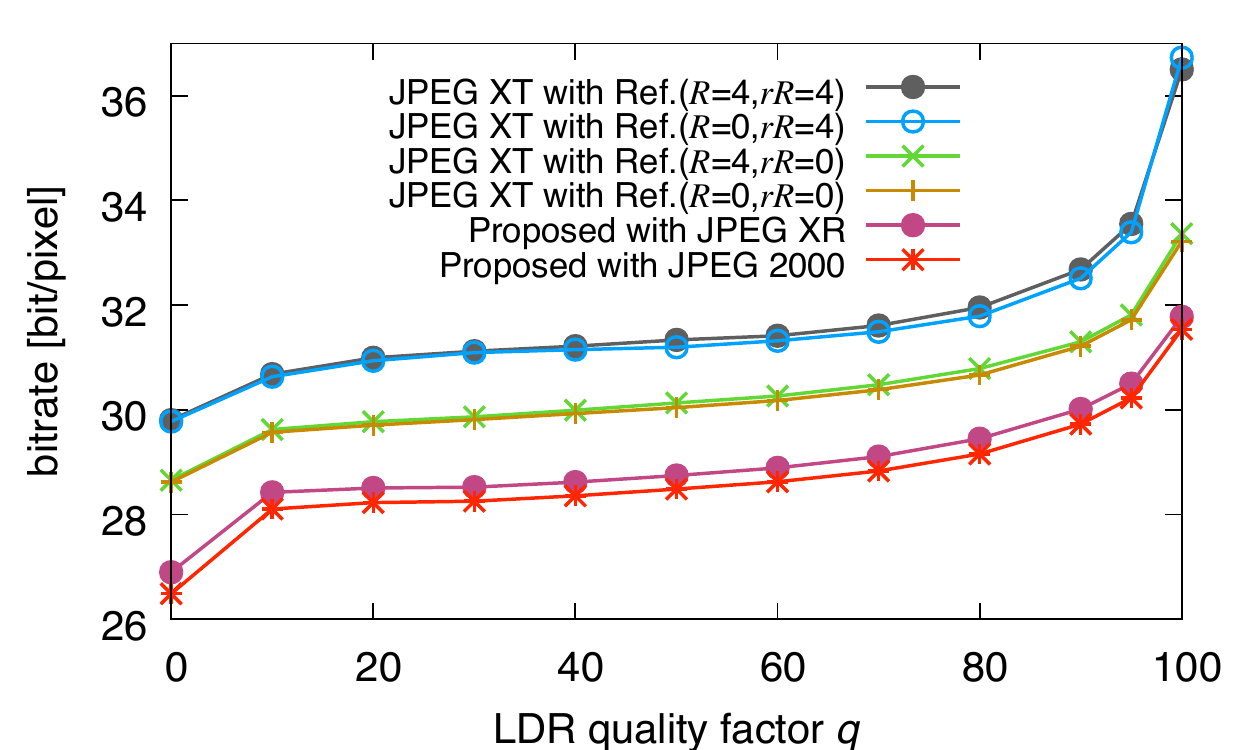}
  }
  \subfigure[i4\label{res_i4}]{
  \includegraphics[width=0.40\linewidth]{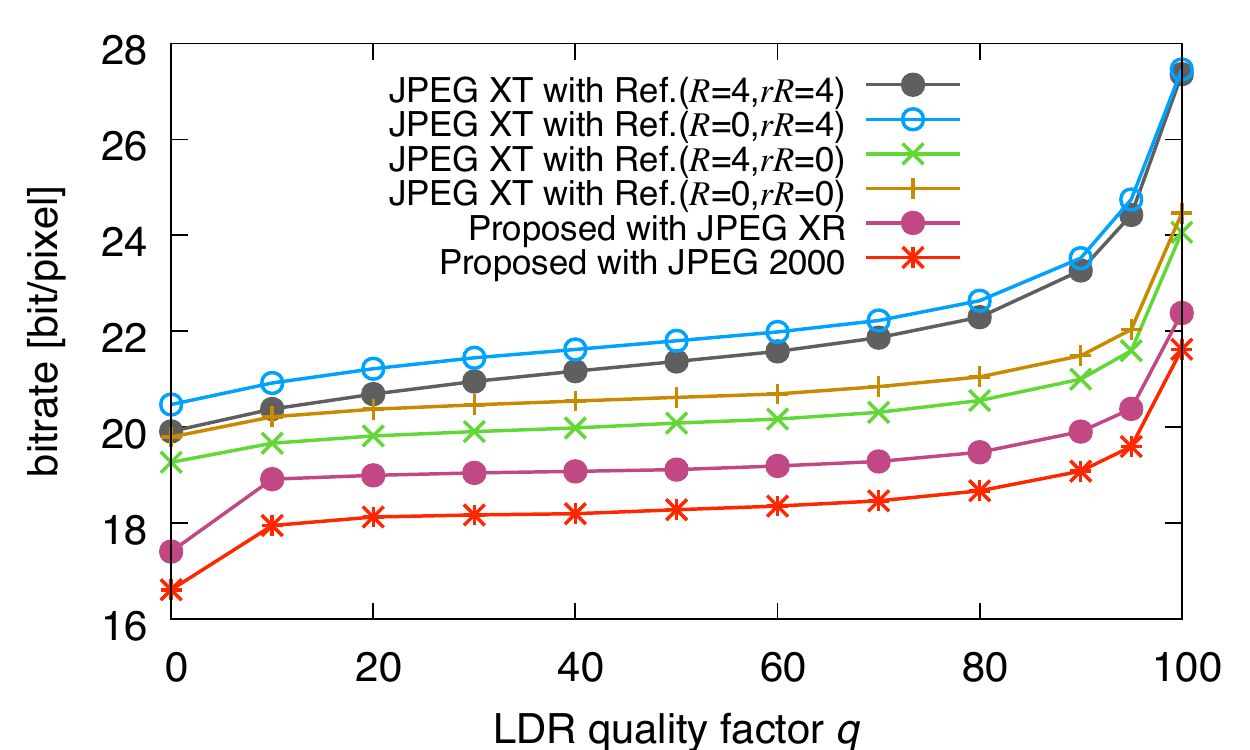}
   }
  \caption{Bitrates of lossless compressed image (integer): image names are represented by index (see Table. \ref{testimages}.)}
  \label{integer_results}
  \end{figure*}

\begin{figure}[tb]
    \centering
    \subfigure[Floating-point\label{unpack_f}]{
    \includegraphics[width=0.8\linewidth]{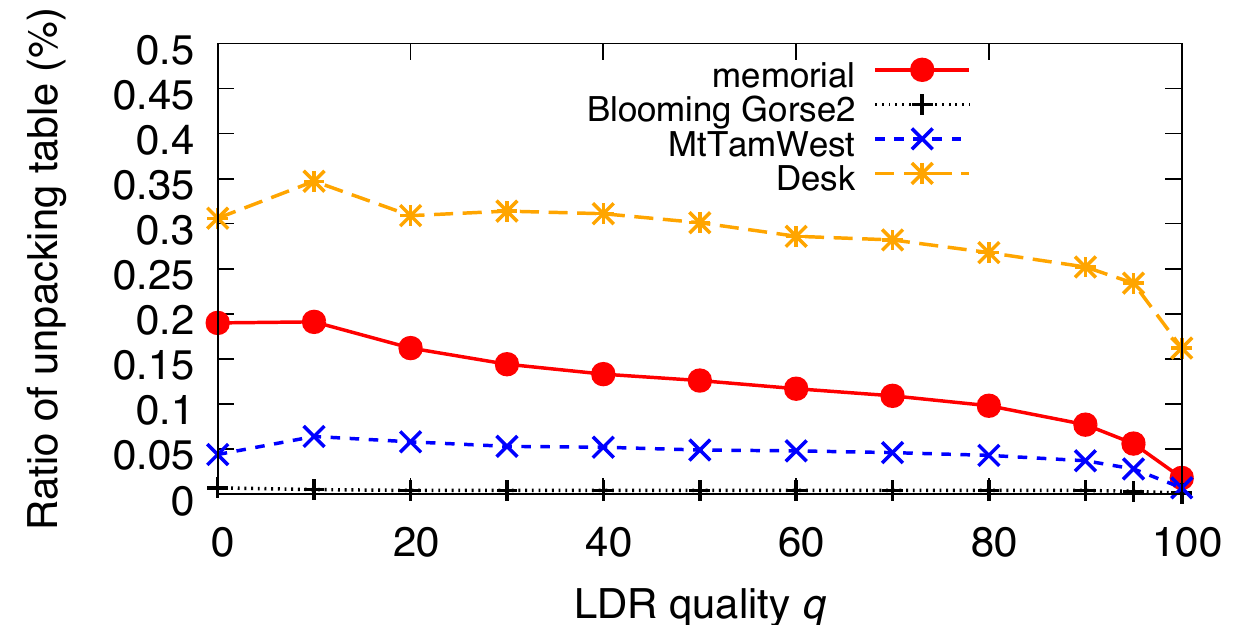}
    }
    \subfigure[Integer\label{unpack_i}]{
    \includegraphics[width=0.8\linewidth]{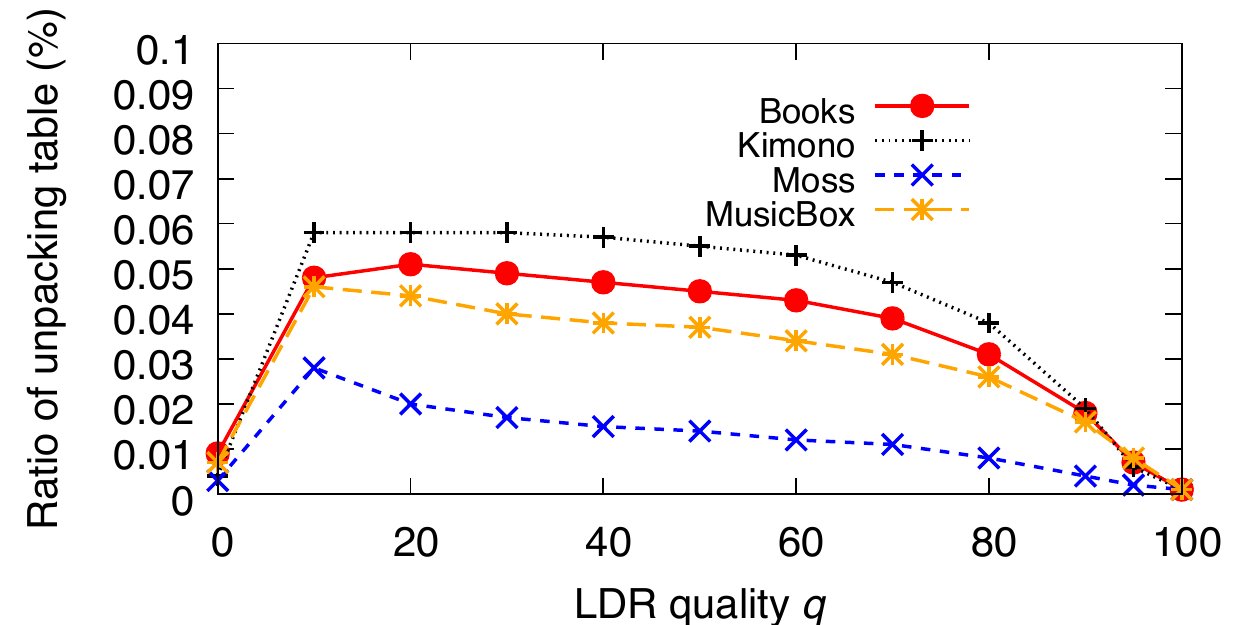}
    }
    \caption{Ratio of unpacking table to total bitrate (\%)}
    \label{ratio_unpack}
   \end{figure}

Figures \ref{floating_results} and \ref{integer_results} show the results of lossless bitrate with the
proposed method with the different LDR quality $q$ and those with the
\myhl{JPEG XT Part 8}{D27} with the different parameter values of
$q$, $R$ and $rR$.
From these results, it is clearly confirmed that the
results of the proposed method show the better lossless performance regardless of
images and those pixel value types, the values of LDR quality $q$. 
It is worth noting that the values of $R$ and/or $rR$ should be
carefully determined for the JPEG XT.
\newsavebox{\boxi}\sbox{\boxi}{\ref{res_f2}}
\myhl{
For example, in Fig. {\usebox\boxi}, the
bitrates of JPEG XT with the combination of the refinement parameters
$(R=4, rR=0)$ illustrated in a green line are higher than 32 bpp from
$q=0$ to $q=60$, while those with the other combinations are lower than
32 bpp. However, the green line is the lowest position over $q=80$.
Thus, the combination $(R=4, rR=0)$ results in the worst
lossless performance between $q=0$ and $q=60$, although it results in the
best performance over $q=80$. 
From these figures, it is observed that the best combination of $R$ and $rR$ depends on the LDR quality $q$ and
the input image.}{B4}
On the other hand, the proposed method with the JPEG 2000 encoder gives
\myhl{the best performance.}{D26}
The second best is the result of proposed method with the
JPEG XR encoder. Although there is some difference between the results,
those two types of the proposed method give the lower bitrate than those
obtained by the JPEG XT encoder, even though there is no dependency on the LDR
$q$ and the input image.
The ratios of the data amount for the unpacking table to the total
bitrate are illustrated in Fig. \ref{ratio_unpack}. The ratio can be
considered to be almost negligible because it is less than 0.4\% at maximum.
\section{Conclusions}
 A novel method using the histogram packing technique with the two-layer coding having the
 backward compatibility with the legacy JPEG for base layer has been proposed
 in this paper.
 The histogram packing technique has been used to improve the performance of
 lossless compression for HDR images that have the histogram sparseness.
 The experimental results in terms of lossless bitrate have demonstrated that the proposed
 method has a higher compression performance than that of the \myhl{JPEG XT Part 8}{D27}. 
\myhl{Unlike the JPEG XT Part 8, there is no need to determine
 image-dependent values of the coding parameters to achieve good compression
 performance. Moreover, as well as the JPEG XT Part 8, the base layer produced by the proposed method preserves the backward compatibility to the legacy JPEG standard.}{B5,D27}

\bibliographystyle{ieicetr}
\bibliography{./bibs/IEEEabrv,./bibs/refs}
\label{sec:ref}
\newpage
\profile{Osamu WATANABE}{%
received his B.S., M.S., and Ph.D. degrees from Tokyo Metropolitan
 University in 1999, 2001, and 2004, respectively. In 2004, He joined
 the faculty of Takushoku University as a research associate in
 electronics and computer systems. Since 2008, He has been an Associate
 Professor in the Department of Electronics and Computer Systems. He is
 a member of IEEE, ITE and an Associate Editor for The Journal of ITE.
He is also an expert of ISO/IEC/JTC1/SG29/WG1 committee “Joint
 Photographic Experts Group” (JPEG). His areas of research interest are
 image processing, image coding, and signal processing.
 }%

\profile{Hiroyuki KOBAYASHI}{%
received his B.E., M.E., and Ph.D. degrees in Electrical Engineering from Tokyo Metropolitan University in 1992, 1994, and 1997.
In 1997, he joined Tokyo Metropolitan College of Technology, where he is currently a Professor in the Electrical and Electronic Engineering Course, Tokyo Metropolitan College of Industrial Technology.
His research interests are in the areas of digital signal processing,
multi-rate systems, and image compression.
}%

\profile{Hitoshi KIYA}{%
 received his B.Eng. and M.Eng. degrees from Nagaoka University of
 Technology, Japan, in 1980 and 1982, respectively, and his
 D.Eng. degree from Tokyo Metropolitan University in 1987. In 1982, he
 joined Tokyo Metropolitan University as an Assistant Professor, where
 he became a Full Professor in 2000. From 1995 to 1996, he attended
 the University of Sydney, Australia as a Visiting Fellow.
 He currently serves as the President of
 APSIPA and Regional Director-at-Large for Region 10 of IEEE Signal
 processing Society.
 He was
 the Chair of IEEE Signal Processing Society Japan Chapter, an Associate
 Editor for IEEE Trans. Image Processing, IEEE Trans. Signal Processing
 and IEEE Trans. Information Forensics and Security, respectively. He
 also served as the President of IEICE Engineering Sciences Society
 (ESS), the Editor-in-Chief for IEICE ESS Publications, and a Vice
 President of APSIPA.
 He received IEEE ISPACS Best Paper Award in 2016,
 IWAIT Best Paper Award in 2014and 2015, ITE Niwa-Takayanagi Best Paper
 Award in 2012, the Telecommunications Advancement Foundation Award in
 2011, and IEICE Best Paper Award in 2008. He is a Fellow of IEEE, IEICE
 and ITE.
}%

\end{document}